\documentclass[11pt]{article}
\usepackage[margin=1.0in]{geometry}
\usepackage{cite}
\usepackage{amsmath,amssymb,amsfonts}
\usepackage{graphicx}
\usepackage{textcomp}
\usepackage{booktabs}
\usepackage{multirow}
\usepackage{algorithm}
\usepackage{algorithmic}
\usepackage{soul}
\usepackage[caption=false,font=small]{subfig}
\usepackage[hidelinks]{hyperref}

\title{Higher-order domain generalization in magnetic resonance-based assessment of Alzheimer's disease}
\author{Zobia Batool\thanks{Zobia Batool, Huseyin Ozkan, and Erchan Aptoula are with the Faculty of Engineering and Natural Sciences (VPA Lab), Sabanci University, Istanbul, T\"urkiye.}\and
Diala Lteif\thanks{Diala Lteif is with the Department of Computer Science, Boston University, Boston, MA, USA.}\and
Vijaya B. Kolachalama\thanks{Vijaya B. Kolachalama is with the Department of Medicine, Boston University Chobanian \& Avedisian School of Medicine, and the Department of Computer Science and Faculty of Computing and Data Sciences, Boston University, Boston, MA, USA.}\and
Huseyin Ozkan\and Erchan Aptoula}
\date{}

\begin{document}
\maketitle

\begin{abstract}
Despite progress in deep learning for Alzheimer's disease (AD) diagnostics, models trained on structural magnetic resonance imaging (sMRI) often do not perform well when applied to new cohorts due to domain shifts from varying scanners, protocols and patient demographics. AD, the primary driver of dementia, manifests through progressive cognitive and neuroanatomical changes like atrophy and ventricular expansion, making robust, generalizable classification essential for real-world use. While convolutional neural networks and transformers have advanced feature extraction via attention and fusion techniques, single-domain generalization (SDG) remains underexplored yet critical, given the fragmented nature of AD datasets. To bridge this gap, we propose Extended MixStyle (EM), a higher-order extension of MixStyle that incorporates skewness and kurtosis into feature-statistic perturbation to mimic diverse distributional variations. Trained on sMRI data from the National Alzheimer's Coordinating Center (NACC; $n=4,647$) to differentiate participants with normal cognition (NC) from those with mild cognitive impairment (MCI) or AD and tested on three unseen cohorts (total $n=3,126$), EM yields enhanced cross-domain performance, with an average numerical gain of 2.4 percentage points in macro-F1 over the 3D U-Net baseline, underscoring its potential for improving cross-cohort generalization in sMRI-based AD classification. The source code is available at \url{https://github.com/zobia111/Extended-Mixstyle}.
\end{abstract}

\noindent\textbf{Keywords:} Alzheimer's disease, domain shift, higher-order statistics, MixStyle extension, single-domain generalization, structural MRI

\section{Introduction}
\label{sec:introduction}
Alzheimer’s disease (AD) is a progressive neurodegenerative disorder and the leading cause of dementia worldwide. Its onset and progression are influenced by aging, genetic predisposition and environmental factors, and are clinically characterized by memory loss, cognitive decline, and behavioral changes \cite{paper1}. Structural magnetic resonance imaging (sMRI) provides a visual presentation of disease-related neuroanatomical changes, including cortical thinning, ventricular enlargement and regional gray matter atrophy.

Deep learning has emerged as a powerful approach for detecting AD-related changes in sMRI. Convolutional neural networks (CNNs) remain the most widely used models, with enhancements such as attention mechanisms, multi-scale feature fusion and multimodal integration to improve sensitivity to subtle morphometric patterns \cite{jabason2024lightweight, wu2022attention, sharma2022conv}. Recent advances, including spatial and channel attention, frequency filtering and tailored optimization strategies, have further improved CNN performance \cite{li2024multi, zhang20223d, sathyabhama2025effective}. In parallel, transformer-based architectures have also gained momentum, leveraging self-attention to capture long-range anatomical dependencies and integrate sMRI with other imaging modalities \cite{joy2025vitad, duong2025multimodal, kunanbayev2024training}. Despite these advances, sMRI data collected across sites and studies vary widely due to differences in scanner manufacturers, acquisition parameters, preprocessing pipelines and participant demographics. This variability gives rise to distributional differences between training and testing data, which often lead models to overfit to cohort-specific patterns rather than disease-specific features \cite{wang2022generalizing, han2023multi, hl2024multimodal, jang2022m3t}. Consequently, high-performing models trained within a single dataset frequently fail to generalize to unseen cohorts. This challenge has motivated growing interest in domain generalization (DG) methods that aim to learn invariant representations robust to such shifts.

Most prior efforts in DG rely on access to multiple labeled domains during training to learn shared invariant features. However, in practice, especially within AD research, large-scale multi-domain training data are rarely available due to privacy constraints and cohort heterogeneity. As a result, the single-domain generalization (SDG) setting, where models must generalize to unseen domains after training on only one dataset, offers a more realistic yet underexplored paradigm. In the AD context, SDG is particularly pertinent: most cohorts are modest in size and independently curated, making aggregation across institutions infeasible. Thus, SDG methods must capture intrinsic variability within a single cohort to prepare models for unknown, real-world data distributions.

\begin{figure}[t]
  \begin{center}
    \subfloat[Mean Skewness $\pm 95$\% CI]{%
        \includegraphics[width=0.47\textwidth]{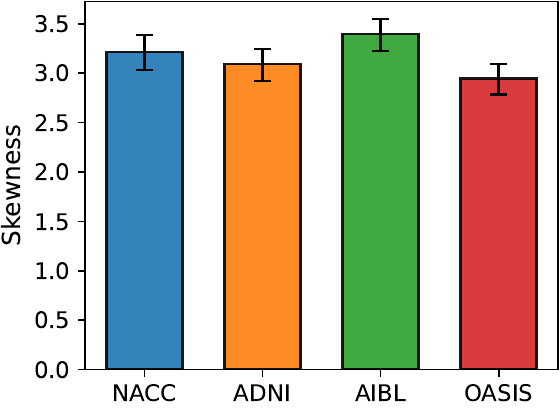}%
        \label{fig:skewness_bar}}
    \hfill
    \subfloat[Mean Kurtosis $\pm 95$\% CI]{%
        \includegraphics[width=0.47\textwidth]{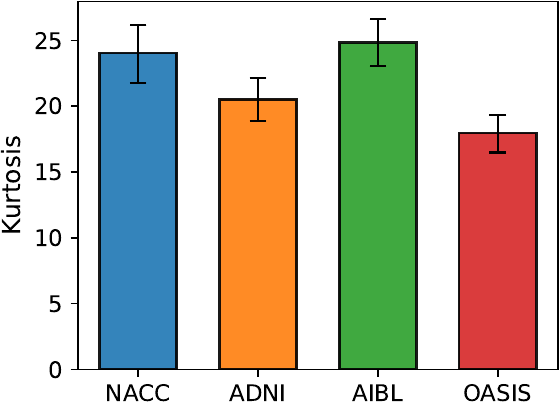}%
        \label{fig:kurtosis_bar}}
    \caption{\textbf{Higher-order feature statistics across cohorts.} Bar plots of skewness and kurtosis computed from intermediate 3D U-Net feature maps across four sMRI cohorts. For each dataset, channel-wise feature distributions were aggregated over test samples and their higher-order moments were summarized. Error bars denote the 95\% confidence intervals (CIs). Clear variability in skewness and kurtosis was observed across cohorts, indicating that feature distributions differ beyond mean and standard deviation.}
    \label{fig:skew_kurt_plots}
  \end{center}
\end{figure}

Recent work on SDG has explored diverse strategies, including patch-free 3D ResNets with domain-specific classifiers informed by similarity metrics \cite{zhou2023learning}, attention-supervised 3D U-Nets guided by SHAP-based saliency priors \cite{lteif2024disease}, and prototype-based alignment coupled with adversarial discriminators \cite{cai2023prototype}. Augmentation-based strategies, such as MixStyle \cite{mixstyle}, have also been explored to improve robustness by perturbing feature statistics during training. Yet, these approaches primarily manipulate first- and second-order moments (mean and standard deviation), which insufficiently capture the richer distributional complexity of 3D sMRI data. Higher-order statistics such as skewness and kurtosis vary significantly across sMRI datasets and reflect subtle, clinically relevant sources of heterogeneity (Fig.~\ref{fig:skew_kurt_plots}). This observation motivates incorporating higher-order moments into feature-statistic perturbations to better simulate domain shifts and improve cross-cohort generalization in AD classification.

To address this challenge, we propose an extension of MixStyle tailored for SDG in sMRI-based AD classification. The proposed extension perturbs intermediate feature maps within a 3D U-Net backbone by blending not only first- but also higher-order moments, specifically skewness and kurtosis. By enriching feature-level perturbations in this manner, the method simulates diverse inter-cohort distributional variations, thereby encouraging domain-invariant representations while preserving sensitivity to AD-related morphometric alterations. To evaluate this approach, the model was trained on sMRI data from the National Alzheimer's Coordinating Center (NACC) \cite{NACC} to distinguish normal cognition (NC), mild cognitive impairment (MCI), and AD. It was tested on three independent cohorts Alzheimer's Disease Neuroimaging Initiative (ADNI)\cite{ADNI}, Australian Imaging Biomarkers and Lifestyle Study of Ageing (AIBL) \cite{AIBL} and Open Access Series of Imaging Studies (OASIS-3) \cite{OASIS} with distinct imaging protocols and demographic profiles, to assess out-of-distribution generalization.

\section{Related Work}
\subsection{Classification models on neuroimaging data}
Various approaches have been proposed for classification of neuroimaging data. For instance, one study integrated adaptive skull stripping and handcrafted feature selection with a modified SqueezeNet for efficient classification \cite{sathyabhama2025effective}. Other CNN-based approaches including lightweight hybrid attention networks \cite{jabason2024lightweight} and attention-based multi-scale 3D CNNs have also been investigated \cite{wu2022attention}. Multimodal frameworks have further combined MRI with other modalities: for instance, wavelet-transformed MRI and PET features were combined with CNNs and ensemble random vector functional link classifiers \cite{sharma2022conv}. Other CNN variants include an anatomy-aware gating mechanism \cite{jiang2024anatomy}, Fourier-transform-based 3D networks such as Global Fourier Network \cite{zhang20223d}, and spatial/channel attention modules incorporated into 3D ResNet backbones \cite{li2024multi}. Building on CNNs, transformer-based models have also gained popularity. A ViT framework enhanced with Laplacian sharpening was proposed in \cite{joy2025vitad}, while multimodal transformers fused MRI and PET through cross-attention \cite{duong2025multimodal}. Other innovations include synthetic data generation, masked autoencoders, and knowledge distillation to improve performance under limited labeling \cite{seyfiouglu2022brain, kunanbayev2024training}. Beyond CNNs and transformers, graph-based and hybrid models include disentangled attention graph neural networks \cite{gamgam2024disentangled} and lightweight dense attention networks \cite{gan2025dense}. Together, these approaches illustrate the diversity of deep learning strategies applied to AD classification. However, despite these advances, generalization to unseen datasets with different acquisition characteristics remains a critical challenge.

\subsection{Domain generalization frameworks}
Several DG techniques originally developed for general computer vision tasks were adopted into medical imaging classification pipelines. MixUp \cite{zhang2018mixup} focused on interpolating inputs and labels to generate synthetic samples, while MixStyle \cite{mixstyle} perturbs feature statistics by mixing mean and standard deviation across instances, although such randomness can distort disease-relevant features. Alternative augmentations include adversarial Bayesian approaches \cite{cheng2023adversarial}, frequency-based perturbations \cite{li2023frequency}, and extended variants such as random amplitude spectrum synthesis, which incorporates mask reconstruction to further simulate distribution shifts and enhance SDG \cite{rass2025random}. Beyond augmentation, a contrastive SDG method \cite{CCSDG} separated style and structure by using style-augmented image pairs, encouraging segmentation to depend on structure alone. The additive disentanglement of domain features with remix loss \cite{demirel2023adrmx} further advanced disentanglement by subtracting domain features from label features and combining invariant and domain-specific representations to improve robustness. Similarly, gradient-based suppression methods, such as representation self-challenging (RSC) \cite{huangRSC2020}, forced models to leverage alternative cues by masking dominant features. To improve cross-domain feature alignment, exact feature distribution matching (EFDM) \cite{zhang2022exact} replaced Gaussian assumptions with empirical distribution matching via a Sort-Matching algorithm.

More recent medical imaging studies  have further extended these ideas through style-based and adaptive frameworks. For instance, hallucinated style distillation \cite{hsd2025hallucinated} generates diverse styles from a single source and employs cross-domain distillation with a regularization objective to learn style-invariant features. Similarly, CompStyle \cite{spanos2024complex} combines style transfer, adversarial training, and input-level augmentation to mitigate dataset bias. Moreover, gradient-alignment strategies \cite{Ballas_2025_CVPR} have been proposed to mitigate inter-domain conflicts during early training, promoting domain-invariant feature learning and smoother convergence. In addition, parameter-space ensemble method \cite{peer2025pressure} leverages parameter averaging and mutual-information regularization to reduce feature distortion during training, while a dynamic DG framework \cite{cheng2025dynamic} employs position transfer and Fourier transformations to capture style variations. Finally, other optimization-focused approaches \cite{Li_2025_CVPR} refine loss landscapes across domains to achieve consistent flat minima and further enhance robustness.

In the context of DG applied to MRI data, some methods incorporated disease priors, such as region-based interpretability with class-wise attention and saliency maps \cite{lteif2024disease}, or domain-knowledge-constrained CNNs, such as a 3D ResNet with domain-weighted classifiers \cite{zhou2023learning}. Others focused on adversarial strategies, including synthesizers with mutual information regularization \cite{xu2022adversarial}, and self-distillation in vision transformers using softened predictions \cite{galappaththige2024generalizing}. Hybrid frameworks combined multiple strategies to achieve robust performance. For example, a prototype-guided multi-scale domain adaptation \cite{cai2023prototype} combines convolution, attention, and adversarial feature alignment, while domain contrastive learning \cite{fiasam2023domain} integrates contrastive learning into a 3D autoencoder. A recent approach combined a hybrid spatial-channel attention mechanism to refine spatial and channel-wise features with contrastive learning to enforce domain-invariant representations for AD classification across multi-site MRI data \cite{sam2026multisite}. Most recently, structure-aware augmentation methods mixed anatomically coherent regions using distance transforms \cite{batool2025distance}. These contributions collectively demonstrate a growing emphasis on domain generalization as a prerequisite for reliable MRI-based classification. However, existing approaches largely overlook higher-order feature statistics, motivating the extension of MixStyle to capture richer distributional variations for improved cross-domain generalization.

\section{Extended MixStyle}
\begin{figure}[!t]
    \begin{center} \includegraphics[width=\textwidth]{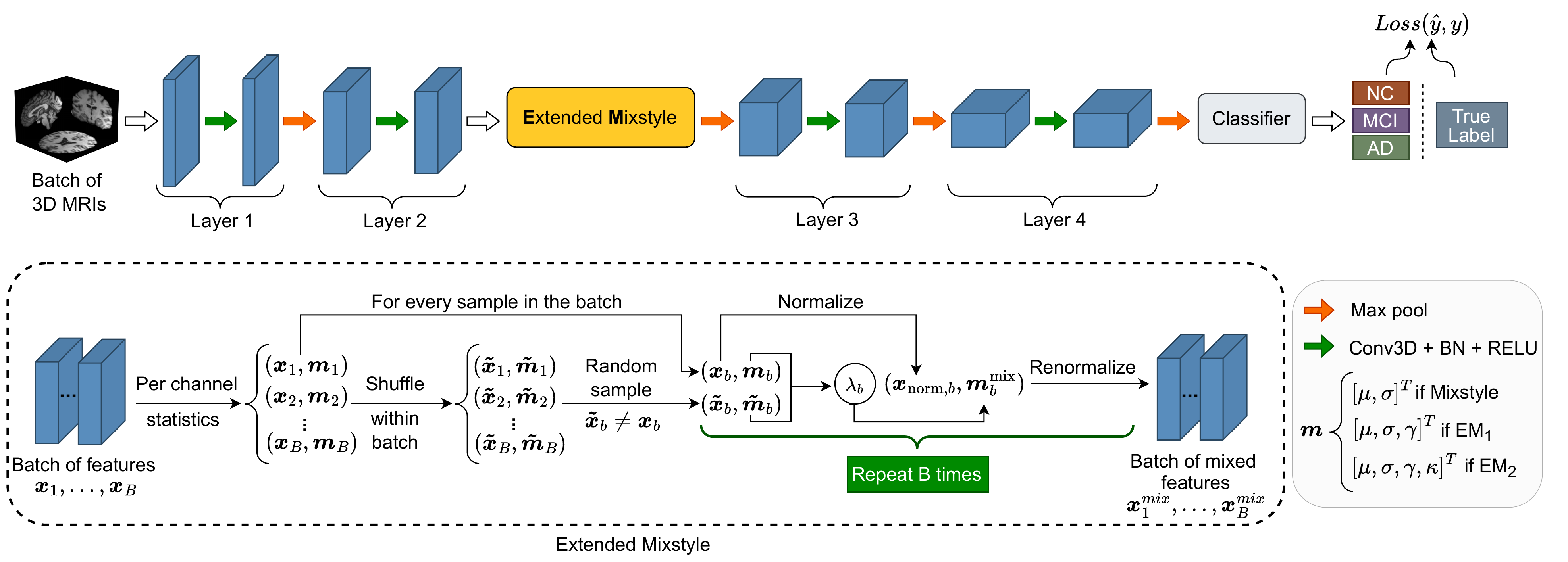}
        \caption{\textbf{\text{EM} integration into a 3D U-Net encoder and its internal operation.} EM receives a batch of feature maps as input, and produces a batch of mixed feature maps as output. In detail, the batch $(\boldsymbol{x}_1, \boldsymbol{x}_2, \ldots, \boldsymbol{x}_B)$ are obtained from Layer 2, and are used to compute per-channel feature statistics $m$.
        For each sample $b$, the mean ($\mu_b$), standard deviation ($\sigma_b$), skewness ($\gamma_b$), and  kurtosis ($\kappa_b$) are computed (depending on the specific variant, $EM_1$ or $EM_2$). These statistics are randomly paired (e.g.~between $\boldsymbol{x}_b$ and $\tilde{\boldsymbol{x}}_b$), and mixed through a random sample-specific mixing coefficient $\lambda_b$
        producing mixed statistics $\boldsymbol{m}^{mix}_b$. Each feature map is then normalized using its original statistics and renormalized using its corresponding mixed statistics, thus generating mixed feature maps that are forwarded to the next encoder layer. 
}
\label{fig:mixstyle_pipeline}
    \end{center}
\end{figure}

Given a 3D sMRI volume, the objective is to improve SDG by perturbing intermediate feature distributions during training. To achieve this, an extension to MixStyle \cite{mixstyle} is introduced, which augments the original MixStyle framework by incorporating higher-order statistical moments, specifically skewness and kurtosis, in addition to the conventional mean and standard deviation. Two variants of this module are considered for evaluation. The first variant extends MixStyle with skewness and is referred to as EM\textsubscript{1}. The second variant extends MixStyle with both skewness and kurtosis and is referred to as EM\textsubscript{2}. These modules are integrated into a 3D U-Net backbone, where each sMRI volume is processed through the network with Extended MixStyle (EM) applied at selected layers during training to encourage domain-invariant representation learning. Designed to operate without access to multi-domain data, the proposed approach aims to enhance the robustness of AD classification across unseen cohorts, as illustrated in Fig.~\ref{fig:mixstyle_pipeline}.

\subsection{Model architecture}
The classification framework is based on a 3D U-Net architecture \cite{Unet3D}, which serves as the backbone for feature extraction. The architecture comprises four stacked convolutional blocks, each consisting of two convolutional layers followed by continuous batch normalization and ReLU activation, with Extended MixStyle regularization applied to the second intermediate layer. This specific placement introduces style variation at a mid-level semantic representation, which is empirically found to have an effective balance between low-level noise and high-level abstraction. To leverage prior knowledge, the model is initialized with pre-trained weights derived from chest CT scans \cite{Unet3D}. For adaptation to the classification task, the decoder component of the U-Net was removed. The resulting high-level feature representations were then globally average-pooled, followed by two fully connected layers to produce the final classification output, as shown in Fig.~\ref{fig:mixstyle_pipeline}.

\subsection{MixStyle framework}
MixStyle \cite{mixstyle} performs feature-level domain mixing by interpolating the statistical moments of feature maps, specifically the mean and standard deviation, across spatial dimensions. 

Given a batch of feature maps $\boldsymbol{x} \in \mathbb{R}^{B \times C \times D \times H \times W} $, where $ B $ is the batch size, $ C $ is the number of channels, $ D $, $ H $, and $ W $ are the depth, height, and width of the feature map respectively, MixStyle first computes the per-channel spatial mean  $\mu(\boldsymbol{x}) \in \mathbb{R}^{B \times C}$ and standard deviation $\sigma(\boldsymbol{x}) \in \mathbb{R}^{B \times C}$:
\begin{equation}
\mu_{b,c}(\boldsymbol{x}) = \frac{1}{N} \sum_{i=1}^{N} x_{b,c,i}
\label{eq:mean}
\end{equation}
\begin{equation}
\sigma_{b,c}(\boldsymbol{x}) = \sqrt{ \frac{1}{N} \sum_{i=1}^{N} (x_{b,c,i} - \mu_{b,c}(\boldsymbol{x}))^2 + \varepsilon }
\label{eq:std}
\end{equation}
\noindent where 
${x}_{b,c,i}$ denotes the scalar activation value at the $i$-th 3D spatial position of the $c$-th channel in the $b$-th sample, and $N = D \times H \times W$ is the total number of spatial positions within each feature map. Moreover, $\varepsilon = 10^{-6}$ is added to ensure numerical stability.

After computing the per-channel statistics for each sample in the batch, a random permutation was applied along the batch dimension so that its statistics $(\mu(\boldsymbol{x}), \sigma(\boldsymbol{x}))$ were paired with another distinct random sample's statistics $(\mu(\tilde{\boldsymbol{x}}), \sigma(\tilde{\boldsymbol{x}}))$, that were then mixed:

\begin{equation}
\mu^{\text{mix}}_{b,c} = \lambda_b \, \mu_{b,c}(\boldsymbol{x}) + (1 - \lambda_b) \, \mu_{b,c}(\tilde{\boldsymbol{x}})
\label{eq:mu_mix}
\end{equation}
\begin{equation}
\sigma^{\text{mix}}_{b,c} = \lambda_b \, \sigma_{b,c}(\boldsymbol{x}) + (1 - \lambda_b) \, \sigma_{b,c}(\tilde{\boldsymbol{x}})
\label{eq:sigma_mix}
\end{equation}
\noindent via a sample-specific mixing coefficient $\lambda_b \sim \text{Beta}(\alpha,  \alpha)$, where \(\lambda_b \in [0, 1]\), and $\alpha > 0$ is the concentration parameter of the Beta distribution regulating the degree of interpolation between them. Each feature map in the batch was then first normalized using its own instance-level statistics across spatial dimensions: 
\begin{equation}
{x}_{\text{norm},b,c,i} = 
\frac{x_{b,c,i} - \mu_{b,c}(\boldsymbol{x})}{\sigma_{b,c}(\boldsymbol{x})}
\label{eq:norm}
\end{equation}
\noindent and then reparameterized (i.e.~rescaled) using the mixed statistics:
\begin{equation}
x^{mix}_{b,c,i} =
{x}_{\text{norm},b,c,i} \cdot \sigma^{\text{mix}}_{b,c} + \mu^{\text{mix}}_{b,c}
\label{eq:mixstyle_final}
\end{equation}
\noindent
Eq.~\eqref{eq:mixstyle_final} preserves the content information of the feature map while blending its style with that of another sample in the batch, hence promoting domain-invariant representations \cite{mixstyle}.

However, although MixStyle addresses first- and second-order shifts, inter-cohort differences in sMRI often involve more complex distributional variations as shown in Fig.~\ref{fig:skew_kurt_plots}. In fact, sMRIs contain complex anatomical structures and non-Gaussian intensity patterns that can influence higher-order moments like skewness and kurtosis \cite{bahsoun2022flair}. 

Motivated by this observation, we propose two extensions to MixStyle, namely $EM_1$ and $EM_2$ (short for Extended MixStyle) that use mixed skewness and kurtosis statistics to guide additional feature-space perturbations beyond the conventional mean- and standard-deviation-based transformation. These higher-order-moment-guided perturbations are intended to improve robustness to inter-domain variability.

\subsection{Extended MixStyle with higher-order moments}
Given the input batch of feature maps $\boldsymbol{x}$, the per-sample, per-channel skewness $\gamma(\boldsymbol{x}) \in \mathbb{R}^{B \times C}$ is computed as:  
\begin{equation}
\gamma(\boldsymbol{x})_{b,c} = \frac{1}{N} \sum_{i=1}^{N} \frac{({x}_{b,c,i} - \mu(\boldsymbol{x})_{b,c})^3}{\sigma(\boldsymbol{x})_{b,c}^3}
\label{eq:skewness}
\end{equation}
The mixed skewness was then obtained through the same interpolation strategy as in MixStyle Eqs.~\eqref{eq:mu_mix} and \eqref{eq:sigma_mix}, 
using the mixing coefficient $\lambda_b$:
\begin{equation}
\gamma^{\text{mix}}_{b,c} = \lambda_b \, \gamma(\boldsymbol{x})_{b,c} + (1 - \lambda_b) \, \gamma(\tilde{\boldsymbol{x}})_{b,c}
\label{eq:gamma_mix}
\end{equation}

To simulate asymmetric, non-Gaussian feature variations, the MixStyle formulation in
Eq.~\eqref{eq:mixstyle_final} was extended by incorporating skewness. This variant, referred to as EM\textsubscript{1}, perturbs feature distributions using the mixed third-order moment as a guidance term. The resulting perturbed feature map becomes:

\begin{equation}
\text{EM}_{1}(\boldsymbol{x})_{b,c,i} =
x^{mix}_{b,c,i}
+
\beta_{\text{skew}} \cdot \gamma^{\text{mix}}_{b,c} \cdot
{x}_{\text{norm},b,c,i}^{3}
\cdot \sigma^{\text{mix}}_{b,c}
\label{eq:em1}
\end{equation}
\noindent where the cubic term ${x}_{\text{norm},b,c,i}^{3}$ captures the asymmetric component of the normalized feature distribution, while $\beta_{\text{skew}} \in \mathbb{R}^+$  is the weighting hyperparameter that controls the strength of the skewness-based perturbation.

EM\textsubscript{2} further incorporates kurtosis $\kappa$, defined as:
\begin{equation}
\kappa(\boldsymbol{x})_{b,c} = \frac{1}{N} \sum_{i=1}^{N} 
\frac{({x}_{b,c,i} - \mu(\boldsymbol{x})_{b,c})^4}{\sigma(\boldsymbol{x})_{b,c}^4} - 3
\label{eq:kurtosis}
\end{equation}
\noindent where the subtraction of 3 normalizes the measure such that normally distributed data results in zero kurtosis, commonly referred to as ``excess kurtosis'' \cite{jensen2010mri}. The mixed kurtosis was then computed as:
\begin{equation}
\kappa^{\text{mix}}_{b,c} = \lambda_b \, \kappa(\boldsymbol{x})_{b,c} + (1 - \lambda_b) \, \kappa(\tilde{\boldsymbol{x}})_{b,c}
\label{eq:kappa_mix}
\end{equation}

Finally, building upon EM\textsubscript{1}, EM\textsubscript{2} further uses the mixed kurtosis to guide an additional quartic perturbation term, extending Eq.~\eqref{eq:em1}: 

\begin{equation}
\begin{split}
\text{EM}_{2}(\boldsymbol{x})_{b,c,i} =\,
&\text{EM}_{1}(\boldsymbol{x})_{b,c,i}
+ \beta_{\text{kurt}} \cdot \kappa^{\text{mix}}_{b,c} \cdot
{x}_{\text{norm},b,c,i}^{4} \cdot \sigma^{\text{mix}}_{b,c}
\end{split}
\label{eq:em2}
\end{equation}
\noindent where the term ${x}_{\text{norm},b,c,i}^{4}$ adjusts the tail behavior of the distribution. However, such higher-order term can destabilize training, so weighting hyperparameter $\beta_{\text{kurt}} \in \mathbb{R}^+$ was used to regulate its influence. This incremental design aimed to progressively model complex distributional shifts, capturing both asymmetric and heavy-tailed variations across domains. As with EM\textsubscript{1}, the resulting feature map is guided by the mixed kurtosis rather than constrained to reproduce it exactly.

\section{Experiments}
The proposed method was evaluated against multiple baseline models to assess its effectiveness in AD classification and cross-dataset generalization.

\subsection{Datasets}
Four publicly available cohorts were employed: the National Alzheimer’s Coordinating Center (NACC)~\cite{NACC}, the Alzheimer’s Disease Neuroimaging Initiative (ADNI)~\cite{ADNI}, the Australian Imaging, Biomarkers, and Lifestyle (AIBL) Study~\cite{AIBL}, and the Open Access Series of Imaging Studies (OASIS-3)~\cite{OASIS}. Each dataset contains 3D sMRI scans categorized into three diagnostic groups: NC, MCI, and dementia due to AD (or simply AD). To account for inter-cohort variations in clinical assessment protocols, diagnostic labels were harmonized across all datasets using standardized Clinical Dementia Rating (CDR) staging~\cite{morris1993clinical}, common neuropsychological test batteries, and cohort-specific data dictionaries following established cross-cohort harmonization protocols in AI-based AD diagnostics~\cite{NACC,ADNI,AIBL,OASIS,lteif2024disease,xue2024ai}. Specifically, primary diagnoses in NACC were derived using the Uniform Data Set (UDS v3) variable \texttt{NACCUDSD} (NC: \texttt{NACCUDSD = 1}, CDR = 0; MCI: \texttt{NACCUDSD = 3}, CDR = 0.5 with cognitive complaints and objective impairments in at least one cognitive domain while preserving functional independence; AD: \texttt{NACCUDSD = 4}, CDR $\ge 1$ with fulfillment of NIA-AA or NINCDS criteria for probable AD).
In ADNI, diagnostic labels were mapped via summary variables \texttt{DXDEF}/\texttt{DXCHANGE}, which classify participants according to ADNI's standardized clinical criteria combining global CDR staging, Mini-Mental State Examination (MMSE) scores, and objective memory performance (NC: CDR = 0, MMSE 24--30, non-depressed and non-demented with no objective cognitive complaints; MCI: CDR = 0.5, MMSE 24--30, preserved functional activities, and objective memory loss defined by education-adjusted scores on the Wechsler Memory Scale-Revised (WMS-R) Logical Memory II subtest~\cite{wechsler1987wms}, without meeting clinical criteria for dementia; AD: CDR $\ge$ 0.5, MMSE 20--26, meeting NINCDS/ADRDA criteria for probable AD).
In AIBL, clinical staging was determined by the AIBL Clinical Panel based on extensive neuropsychological testing and Winblad et al.~\cite{winblad2004mild} (NC: CDR = 0, MMSE 26--30; MCI: CDR = 0.5 with objective memory impairment; AD: CDR $\ge 0.5$, MMSE $\le 26$, fulfilling NINCDS-ADRDA criteria). In OASIS-3, clinical diagnoses were established via the Washington University Knight ADRC Clinical Core protocols~\cite{OASIS} based on CDR staging (NC: CDR = 0; MCI: CDR = 0.5 representing very mild dementia/MCI; AD: CDR $\ge 1$ representing probable AD dementia).

\begin{table}[ht]
\caption{\textbf{Participant demographics across sMRI cohorts.} 3D MRI and demographic data were obtained from NACC, ADNI, AIBL, and OASIS-3. Participants were grouped as NC, MCI, or AD, with mean age and number of male participants reported where available.}

\label{table_demographics}
\begin{center}
\setlength{\tabcolsep}{8pt} 

\scriptsize
\begin{tabular}{l c c c}
\hline
\textbf{Dataset} & \textbf{Group (Participants)} & \textbf{Age, years} & \textbf{Gender} \\
                 &                               & \textbf{mean ± std} & \textbf{male count} \\
\hline
\multirow{3}{*}{NACC \cite{NACC}} & NC (n=2524) & 69.8 ± 9.9  & 871 (34.5\%) \\
                      & MCI (n=1175) & 74.0 ± 8.7  & 555 (47.2\%) \\
                      & AD (n=948)   & 75.0 ± 9.1  & 431 (45.5\%) \\
\hline
\multirow{3}{*}{ADNI \cite{ADNI}} & NC (n=481)  & 74.3 ± 6.0  & 235 (48.9\%) \\
                      & MCI (n=971)  & 72.8 ± 7.7  & 572 (58.9\%) \\
                      & AD (n=369)   & 74.9 ± 7.8  & 203 (55.0\%) \\
\hline
\multirow{3}{*}{AIBL \cite{AIBL}} & NC (n=480)  & 72.5 ± 6.2  & 203 (42.3\%) \\
                      & MCI (n=102)  & 74.7 ± 7.1  & 53 (52.0\%)  \\
                      & AD (n=79)    & 73.3 ± 7.8  & 33 (41.8\%)  \\
\hline
\multirow{3}{*}{OASIS-3 \cite{OASIS}} & NC (n=424)  & NA  & NA\\
                      & MCI (n=27)   & NA & NA \\
                      & AD (n=193)   & NA & NA \\
\hline
\end{tabular}
\end{center}
\end{table}

All sMRI scans and associated clinical data were obtained in de-identified form from established research repositories under the respective data-access and Data Use Agreements. The authors had no access to direct participant identifiers and did not interact with or contact any participants. No new data were collected for this study. Because this work constitutes secondary analysis of existing de-identified data obtained under the applicable Data Use Agreements, it was determined not to constitute human-subjects research and did not require additional institutional review board (IRB) approval at the authors' institutions. No uncertain or unresolved diagnoses were present in the cohorts. As each participant contributed a single 3D MRI scan, the train-validation split was inherently participant-level. The selection criterion included T1-weighted 1.5 and 3 Tesla MRI scans of individuals aged 55 years and older, taken within $\pm$6 months from the date of clinically confirmed diagnosis. Subjects younger than 55 years were excluded to minimize age-related effects. Demographic statistics and diagnostic distributions for all cohorts are provided in Table~\ref{table_demographics}.

All sMRI volumes were preprocessed using a standardized pipeline adapted from~\cite{qiu2022multimodal}. The scans were first reoriented to match the MNI space. Brain extraction was performed using the FSL BET tool \cite{smith2002fast}, generating a mask that preserved gray matter, white matter, cerebrospinal fluid, and subcortical regions, while excluding extracranial tissue, brain stem and cerebellum. Following skull stripping, a two-stage linear registration was applied: an initial affine alignment to the MNI-152 coordinate system, followed by repeated skull stripping and registration to refine alignment and remove residual non-brain voxels. Intensity inhomogeneities were then corrected using N4 bias field correction to reduce artifacts and enhance inter-subject consistency. Despite uniform preprocessing across datasets, inter-dataset differences were observed, likely arising from variations in scanners, imaging protocols, and participant demographics. These differences are reflected in the t-SNE embeddings from the ablation study (see Fig.~\ref{fig:tsne-1}, cf. Section~\ref{sec:ablation}), where representations obtained from the 3D U-Net exhibit dataset-specific clustering patterns, thus making the datasets a strong testbed for SDG.

\subsection{Experimental settings}
All experiments were conducted on an NVIDIA A6000 GPU. Due to hardware constraints, training was performed with an effective batch size of 16, achieved through gradient accumulation with a physical batch size of 2. To address class imbalance, weighted cross-entropy loss was employed, with class weights set inversely proportional to class frequencies (NC, MCI, AD). The model was optimized using stochastic gradient descent with an initial learning rate of 0.01, momentum of 0.9, and weight decay of 0.0005. A learning rate scheduler with exponential decay was applied, reducing the learning rate by 5\% after each epoch. Under this configuration, training converged within 60 epochs.

To evaluate DG performance, the standard SDG protocol described in~\cite{qiao2020learning} was adopted. Model training, validation, and hyperparameter selection were conducted exclusively on the NACC cohort using an 80/20 train--validation split. For each configuration, the checkpoint with the highest macro-F1 score on the NACC validation split was selected. The external cohorts, namely ADNI~\cite{ADNI}, AIBL~\cite{AIBL}, and OASIS-3~\cite{OASIS}, were not used for early stopping, checkpoint selection, hyperparameter tuning, or any model-selection decision. All model specifications were fixed using the NACC training and validation data before evaluation on the external cohorts. After the final model configuration was fixed, ADNI, AIBL, and OASIS-3 were used for the primary out-of-distribution evaluation and additional post-hoc analyses, including ablation, sensitivity, visualization, binary classification, and statistical significance analyses.

The proposed model was benchmarked against a baseline 3D U-Net encoder~\cite{Unet3D} without SDG components and several established SDG techniques, including MixUp~\cite{zhang2018mixup} with $\alpha = 0.3$, RSC~\cite{huangRSC2020} with 20\% feature dropout, 5\% background dropout, and a mixing probability of 0.3, EFDM~\cite{zhang2022exact} with a patch replacement probability of $p = 0.5$ and interpolation factor $\alpha = 0.1$, MixStyle~\cite{mixstyle} with $\alpha = 0.1$ and $p = 0.5$, and CCSDG with Feature Distribution Alignment (FDA)~\cite{CCSDG} ratio $L \sim [0.05, 0.1]$, where $L$ denotes the low-frequency spectrum replacement ratio. In addition, recent comparison methods were also included namely DeTTA~\cite{wen2024denoising}, ViT-LD~\cite{kunanbayev2024training}, AAGN~\cite{jiang2024anatomy}, RASS~\cite{rass2025random} and FM-SDG~\cite{li2023frequency}. For these methods, the default hyperparameters from their publicly available implementations were used. All hyperparameters were tuned on the validation set to ensure fair comparison across methods. Model performance was evaluated using four metrics: accuracy, macro-averaged F1 score, sensitivity, and specificity. For binary classification experiments, F1 refers to the positive-class F1 score, with the positive class defined separately for each diagnostic task.

For the proposed approach, both EM\textsubscript{1} and EM\textsubscript{2} modules were integrated into the second layer of the 3D U-Net based on source-validation experiments conducted on NACC. The subsequent external-cohort ablation (see Table~\ref{Layer_experiments}, cf. Section~\ref{sec:ablation}) was performed after this configuration had been fixed and was used only to examine how module placement affects cross-cohort generalization. In the experiments, placing EM in deeper or shallower layers resulted in reduced performance. However, the optimal integration layer may vary depending on the specific architecture or task characteristics.

During training, the EM module induces style perturbations with probability $p$ and is disabled at test time. Gradients through the computed statistics are detached to ensure EM functions purely as feature-space augmentation rather than a learnable transformation. EM\textsubscript{1}  uses an interpolation parameter of $\alpha = 0.7$ and EM\textsubscript{2} $\alpha = 0.5$, both with a mixing probability of $0.9$. These hyperparameters were determined based on the NACC validation split and fixed for all external evaluations. The sensitivity of EM\textsubscript{1} to different $\alpha$--$p$ configurations is further analyzed through macro-F1 heatmaps (see Fig.~\ref{fig:hyperparameter_sensitivity}, cf. Section~\ref{sec:ablation}). The weighting hyperparameters $\beta_{\text{skew}}$ and $\beta_{\text{kurt}}$, introduced in Eqs.~(\ref{eq:em1}) and (\ref{eq:em2}), were set empirically to $\beta_{\text{skew}} = 0.3$ and $\beta_{\text{kurt}} = 0.1$, which provided stable optimization. Finally, EM\textsubscript{1} and EM\textsubscript{2} retain the same 3D U-Net backbone and learnable parameter count as the baseline (19.6M parameters, 1667.1 GFLOPs, and 78.4 MB), with no additional inference-time modules or model complexity. The higher-order perturbations introduce a moderate training-time overhead, increasing the average per-epoch training time by 29.92\% for EM\textsubscript{1} and 32.76\% for EM\textsubscript{2} relative to the baseline (see Table~\ref{tab:computational_efficiency}, Section~IV-D).

\subsection{Results and discussion}
Generalization results across ADNI, AIBL and OASIS-3 cohorts are summarized in Table~\ref{table_results_all}. On ADNI, EM\textsubscript{1} achieved the best sensitivity and F1 score and tied with DeTTA for the best specificity, surpassing CCSDG by up to 3.1 percentage points, while EM\textsubscript{2} showed slightly lower performance. On AIBL, EM\textsubscript{1} provided the most balanced improvements, outperforming MixUp by 1.4--3.4 percentage points across metrics; although EFDM achieved the highest accuracy, it lagged in sensitivity and F1. On OASIS-3, EM\textsubscript{1} tied with the baseline for the best specificity and with EFDM for the best F1, while EM\textsubscript{2} achieved the best sensitivity with comparable specificity. Unlike competing methods, EM\textsubscript{1} showed comparatively balanced numerical performance across cohorts, supporting the effectiveness of higher-order moment perturbations for domain-invariant representation learning.

\begin{table*}[ht]
\caption{\textbf{Cross-dataset generalization performance on external cohorts.} Models were trained on NACC and evaluated on ADNI, AIBL, and OASIS-3. All values in this table are means across three random seeds (1, 2, and 42); $^{*}$ marks methods for which corresponding 10-seed mean results are reported in Table~\ref{table_repeated_runs_f1}. Best results are in bold.}

\label{table_results_all}
\begin{center}
\setlength{\tabcolsep}{5pt} 
\scriptsize
\begin{tabular}{l|cccc|cccc|cccc}
\hline
\multirow{2}{*}{\textbf{Methods}} & \multicolumn{4}{c|}{\textbf{ADNI}} & \multicolumn{4}{c|}{\textbf{AIBL}} & \multicolumn{4}{c}{\textbf{OASIS-3}} \\
 & \textbf{ACC} & \textbf{SEN} & \textbf{SPE} & \textbf{F1} 
 & \textbf{ACC} & \textbf{SEN} & \textbf{SPE} & \textbf{F1} 
 & \textbf{ACC} & \textbf{SEN} & \textbf{SPE} & \textbf{F1} \\
\hline
Baseline$^{*}$  \cite{Unet3D}              & 49.47 & 0.563 & 0.744 & 0.508  & 70.80 & 0.574 & 0.805 & 0.575  & 66.45 & 0.578 & \textbf{0.844} & 0.534 \\
MixUp \cite{zhang2018mixup}         & 46.34 & 0.548 & 0.732 & 0.476  & 75.03 & 0.593 & 0.821 & 0.595  & 65.21 & 0.562 & 0.838 & 0.523 \\
RSC \cite{huangRSC2020}             & 47.17 & 0.546 & 0.732 & 0.483  & 68.22 & 0.583 & 0.817 & 0.561  & 63.04 & 0.549 & 0.830 & 0.514 \\
CCSDG \cite{CCSDG}                  & 47.39 & 0.570 & 0.740 & 0.488  & 73.22 & 0.578 & 0.820 & 0.573  & 67.54 & 0.587 & 0.840 & 0.539 \\
MixStyle \cite{mixstyle}            & 46.62 & 0.558 & 0.737 & 0.476  & 66.26 & 0.575 & 0.811 & 0.538  & 64.44 & 0.565 & 0.838 & 0.518 \\
EFDM \cite{zhang2022exact}          & 43.93 & 0.546 & 0.732 & 0.447  & \textbf{78.81} & 0.551 & 0.796 & 0.582  & \textbf{71.58} & 0.572 & 0.835 & \textbf{0.540} \\  
DT-Mix \cite{batool2025distance} & 45.08 & 0.543 & 0.729 & 0.463  & 74.43 & 0.589 & 0.818 & 0.593  & 65.52 & 0.557 & 0.832 & 0.518 \\

DeTTA \cite{wen2024denoising}         
& 49.15
& 0.543 & \textbf{0.748}
& 0.507  & 62.48 
& 0.552 & 0.807 
& 0.567  & 57.45 
& 0.571 & 0.822 & 
0.488 \\

ViT-LD \cite{kunanbayev2024training}                      &
32.07 & 0.414
& 0.680 & 0.305 
& 59.76 & 0.402 
& 0.712 & 0.359 
& 64.75 & 0.429 
& 0.739 & 0.430 \\

AAGN \cite{jiang2024anatomy}                           & 48.65 & 0.533
& 0.734
& 0.496 
& 74.28
& 0.556
& 0.810 
& 0.571  
& 66.92 
& 0.570 
& 0.830 
& 0.503 \\

RASS \cite{rass2025random}                         & 48.59
& 0.546   
& 0.732    
&0.502 
& 68.22 & 
0.540   
& 0.792    & 0.553  
& 60.40    & 0.537    
& 0.819    & 0.498    \\

FM-SDG \cite{li2023frequency}     & 48.27  
& 0.545   &
0.731   & 
0.500  & 
68.68 & 
0.539   
&
0.797    
& 0.549 
& 61.80    
& 0.548  
& 0.823  
& 0.507   \\

EM\textsubscript{1}$^{*}$               & \textbf{50.30} & \textbf{0.575} & \textbf{0.748} & \textbf{0.519}  & 76.39 & \textbf{0.614} & \textbf{0.836} & \textbf{0.629}  & 68.32 & 0.588 & \textbf{0.844} & \textbf{0.540} \\
EM\textsubscript{2}$^{*}$               & 49.42 & 0.568 & 0.742 & 0.508  & 66.71 & 0.613 & 0.822 & 0.560  & 64.75 & \textbf{0.601} & 0.840 & 0.538 \\
\hline
\end{tabular}
\end{center}
\end{table*}

To assess training stability, the core comparison between the baseline 3D U-Net, EM\textsubscript{1}, and EM\textsubscript{2} was repeated across 10 independent runs using random seeds as shown in Table~\ref{table_repeated_runs_f1}. Statistical significance of the principal macro-F1 comparisons was then evaluated using paired stratified bootstrap resampling at the patient level with 10,000 bootstrap resamples drawn with replacement \cite{efron1992bootstrap}. The same resampled patients were used across methods, and macro-F1 was averaged across the 10 random seeds before computing pairwise differences. Two-sided 95\% confidence intervals and $p$-values were reported, with statistical significance defined at $p \leq 0.05$. Two-sided $p$-values were estimated as twice the smaller of the two empirical bootstrap tail probabilities. Holm correction was applied to the three pairwise comparisons within each external cohort \cite{holm1979simple}.

\begin{table*}[t]
\caption{\textbf{Macro-F1 scores across 10 repeated runs.} Individual seed results and mean $\pm$ standard deviation are reported for the 3D U-Net baseline, EM\textsubscript{1}, and EM\textsubscript{2} across external cohorts.}
\label{table_repeated_runs_f1}

\begin{center}

\setlength{\tabcolsep}{6pt}
\scriptsize
\begin{tabular}{ll|cccccccccc|c}
\hline

\textbf{Dataset} &
\textbf{Method} &
\textbf{1} &
\textbf{2} &
\textbf{3} &
\textbf{4} &
\textbf{5} &
\textbf{6} &
\textbf{7} &
\textbf{8} &
\textbf{9} &
\textbf{42} &
\textbf{$\mu\pm\sigma$} \\

\hline

\multirow{3}{*}{ADNI}
& Baseline
& 0.523 & 0.516 & 0.477 & 0.444 & 0.476
& 0.491 & 0.485 & 0.480 & 0.490 & 0.485
& 0.487$\pm$0.02 \\

& EM\textsubscript{1}
& 0.508 & 0.528 & 0.482 & 0.516 & 0.510
& 0.520 & 0.516 & 0.513 & 0.520 & 0.521
& 0.513$\pm$0.01 \\

& EM\textsubscript{2}
& 0.509 & 0.517 & 0.501 & 0.485 & 0.497
& 0.504 & 0.501 & 0.499 & 0.503 & 0.497
& 0.501$\pm$0.01 \\

\hline

\multirow{3}{*}{AIBL}
& Baseline
& 0.558 & 0.597 & 0.525 & 0.527 & 0.549
& 0.564 & 0.558 & 0.554 & 0.563 & 0.571
& 0.557$\pm$0.02 \\

& EM\textsubscript{1}
& 0.638 & 0.643 & 0.587 & 0.590 & 0.597
& 0.612 & 0.606 & 0.602 & 0.611 & 0.606
& 0.609$\pm$0.02 \\

& EM\textsubscript{2}
& 0.560 & 0.540 & 0.500 & 0.538 & 0.531
& 0.546 & 0.540 & 0.536 & 0.545 & 0.580
& 0.542$\pm$0.02 \\

\hline

\multirow{3}{*}{OASIS-3}
& Baseline
& 0.546 & 0.528 & 0.461 & 0.463 & 0.519
& 0.534 & 0.528 & 0.461 & 0.483 & 0.528
& 0.505$\pm$0.03 \\

& EM\textsubscript{1}
& 0.529 & 0.549 & 0.539 & 0.536 & 0.536
& 0.541 & 0.539 & 0.538 & 0.540 & 0.542
& 0.539$\pm$0.01 \\

& EM\textsubscript{2}
& 0.531 & 0.560 & 0.453 & 0.487 & 0.513
& 0.528 & 0.522 & 0.518 & 0.527 & 0.522
& 0.516$\pm$0.03 \\

\hline
\end{tabular}

\end{center}
\end{table*}

\begin{table*}[ht]
\caption{\textbf{Statistical significance analysis of F1 across 10 random seeds.} Paired stratified bootstrap testing with 10,000 resamples was used. Pairwise differences are reported with 95\% confidence intervals, raw $p$-values, and Holm-adjusted $p$-values. Bold values indicate comparisons that remain statistically significant after Holm correction ($p<0.05$).}
\label{table_bootstrap_significance}

\begin{center}

\setlength{\tabcolsep}{6pt}
\scriptsize
\begin{tabular}{llcccccc}
\hline

\textbf{Dataset}
& \textbf{Comparison}
& \textbf{F1$_1$}
& \textbf{F1$_2$}
& \textbf{$\Delta$F1}
& \textbf{95\% CI}
& \textbf{Raw $p$}
& \textbf{Holm $p$} \\

\hline

\multirow{3}{*}{ADNI}
& EM\textsubscript{1} vs EM\textsubscript{2}
& 0.5134 & 0.5012 & 0.0122
& [-0.0061, 0.0304]
& 0.1952 & 0.2020 \\

& EM\textsubscript{1} vs Baseline
& 0.5134 & 0.4865 & 0.0269
& [0.0051, 0.0492]
& 0.0152 & \textbf{0.0456} \\

& EM\textsubscript{2} vs Baseline
& 0.5012 & 0.4865 & 0.0147
& [-0.0027, 0.0322]
& 0.1010 & 0.2020 \\

\hline

\multirow{3}{*}{AIBL}
& EM\textsubscript{1} vs EM\textsubscript{2}
& 0.6090 & 0.5417 & 0.0673
& [0.0309, 0.1039]
& 0.0002 & \textbf{0.0006} \\

& EM\textsubscript{1} vs Baseline
& 0.6090 & 0.5566 & 0.0525
& [0.0119, 0.0927]
& 0.0126 & \textbf{0.0252} \\

& EM\textsubscript{2} vs Baseline
& 0.5417 & 0.5566 & -0.0149
& [-0.0454, 0.0161]
& 0.3450 & 0.3450 \\

\hline

\multirow{3}{*}{OASIS-3}
& EM\textsubscript{1} vs EM\textsubscript{2}
& 0.5389 & 0.5161 & 0.0228
& [-0.0042, 0.0500]
& 0.0992 & 0.1984 \\

& EM\textsubscript{1} vs Baseline
& 0.5389 & 0.5051 & 0.0338
& [0.0042, 0.0628]
& 0.0228 & 0.0684 \\

& EM\textsubscript{2} vs Baseline
& 0.5161 & 0.5051 & 0.0110
& [-0.0115, 0.0341]
& 0.3374 & 0.3374 \\

\hline
\end{tabular}

\end{center}
\end{table*}

\begin{table*}[ht]
\caption{\textbf{Effect of EM placement within the encoder on model generalization performance.} The proposed modules EM$_1$ and EM$_2$ were inserted after different encoder blocks of the 3D U-Net. Each configuration corresponds to EM modules applied after individual layers or combinations of layers within the encoder. Best results are in bold.}

\label{Layer_experiments}
\begin{center}
\setlength{\tabcolsep}{5pt}
\scriptsize
\begin{tabular}{ll|cccc|cccc|cccc}
\hline
\multirow{2}{*}{\textbf{Layers}} 
& \multirow{2}{*}{\textbf{EM}} 
& \multicolumn{4}{c|}{\textbf{ADNI}} 
& \multicolumn{4}{c|}{\textbf{AIBL}} 
& \multicolumn{4}{c}{\textbf{OASIS-3}} \\
& 
& \textbf{ACC} & \textbf{SEN} & \textbf{SPE} & \textbf{F1}
& \textbf{ACC} & \textbf{SEN} & \textbf{SPE} & \textbf{F1}
& \textbf{ACC} & \textbf{SEN} & \textbf{SPE} & \textbf{F1} \\
\hline

\multirow{2}{*}{Layer 1}   
& EM\textsubscript{1} & 50.90 & 0.562 & 0.745 & 0.523   & 70.49 & 0.595 & 0.822 & 0.586   & 62.42 & 0.599 & 0.829 & 0.512 \\
& EM\textsubscript{2} & 50.79 & 0.552 & 0.740 & 0.519   & 72.31 & 0.613 & 0.832 & 0.612   & 60.24 & 0.550 & 0.823 & 0.490 \\
\hline

\multirow{2}{*}{Layer 2}     
& EM\textsubscript{1} & 50.30 & \textbf{0.575} & \textbf{0.748} & 0.519   & \textbf{76.39} & \textbf{0.614} & \textbf{0.836} & \textbf{0.629}   & \textbf{68.32} & 0.588 & \textbf{0.844} & \textbf{0.540} \\
& EM\textsubscript{2} & 49.42 & 0.568 & 0.742 & 0.508   & 66.71 & 0.613 & 0.822 & 0.560   & 64.75 & \textbf{0.601} & 0.840 & 0.538 \\
\hline

\multirow{2}{*}{Layer 3}     
& EM\textsubscript{1} & 52.16 & 0.542 & 0.738 & 0.528   & 60.51 & 0.551 & 0.796 & 0.537   & 53.88 & 0.558 & 0.813 & 0.476 \\
& EM\textsubscript{2} & 51.94 & 0.560 & 0.744 & \textbf{0.530}   & 67.17 & 0.594 & 0.816 & 0.575   & 58.38 & 0.537 & 0.822 & 0.494 \\
\hline

\multirow{2}{*}{Layer (1,2)} 
& EM\textsubscript{1} & 50.85 & 0.536 & 0.735 & 0.515   & 68.53 & 0.562 & 0.811 & 0.564   & 57.60 & 0.549 & 0.818 & 0.484 \\
& EM\textsubscript{2} & 52.93 & 0.547 & 0.742 & 0.528   & 57.94 & 0.594 & 0.793 & 0.545   & 51.08 & 0.508 & 0.801 & 0.456 \\
\hline

\multirow{2}{*}{Layer (2,3)} 
& EM\textsubscript{1} & \textbf{54.91} & 0.529 & 0.737 & 0.528   & 45.53 & 0.494 & 0.753 & 0.450   & 42.54 & 0.542 & 0.787 & 0.416 \\
& EM\textsubscript{2} & 51.29 & 0.519 & 0.730 & 0.508   & 67.01 & 0.587 & 0.827 & 0.582   & 54.19 & 0.483 & 0.805 & 0.439 \\
\hline

\multirow{2}{*}{Layer (1,3)} 
& EM\textsubscript{1} & 49.58 & 0.523 & 0.727 & 0.502   & 67.17 & 0.598 & 0.813 & 0.589   & 57.14 & 0.514 & 0.815 & 0.478 \\
& EM\textsubscript{2} & 53.21 & 0.529 & 0.734 & 0.527   & 55.06 & 0.537 & 0.778 & 0.511   & 49.53 & 0.511 & 0.796 & 0.441 \\
\hline
\end{tabular}

\end{center}
\end{table*}

The statistical comparisons in Table~\ref{table_bootstrap_significance} further support the cross-cohort results. EM\textsubscript{1} significantly outperformed the baseline on ADNI ($p_{\mathrm{Holm}}=0.0456$) and AIBL ($p_{\mathrm{Holm}}=0.0252$), and also significantly outperformed EM\textsubscript{2} on AIBL ($p_{\mathrm{Holm}}=0.0006$). On OASIS-3, EM\textsubscript{1} showed a positive difference over the baseline, with a 95\% confidence interval excluding zero and a significant unadjusted $p$-value ($p=0.0228$); however, the comparison did not remain significant after Holm correction ($p_{\mathrm{Holm}}=0.0684$). The OASIS-3 improvement is therefore interpreted as suggestive, but not statistically established after multiple-comparison correction. This result should also be considered in light of the pronounced class imbalance in OASIS-3, particularly the limited MCI group (NC: $n=424$, MCI: $n=27$, AD: $n=193$), which increases uncertainty in F1 estimates.

Although the absolute gains are moderate, they are meaningful in the SDG setting because the model is trained on a single source cohort and evaluated on three independent external cohorts without fine-tuning. In clinical deployment settings, models may encounter variations in scanners, acquisition protocols, and patient populations that differ from those used during development. While prospective clinical evaluation is still required, the observed improvements suggest potential benefits in handling such inter-site variability. EM\textsubscript{1} is considered the recommended default variant because it provides the most consistent cross-cohort generalization. EM\textsubscript{2} is retained as an analytical extension to evaluate the effect of adding kurtosis; however, its weaker performance on AIBL highlights a limitation of the current approach, as the fourth-order perturbation may be more sensitive to cohort-specific feature distributions. For future applications, EM\textsubscript{1} can therefore be used as the default variant when no dataset-specific evidence is available. EM\textsubscript{2} may be considered when direct validation on data representative of the intended application indicates a consistent advantage over EM\textsubscript{1}.

\subsection{Ablation Study}
\label{sec:ablation}

The following ablation and sensitivity experiments were conducted as post-hoc analyses after the final model configuration and hyperparameters had been fixed using the NACC validation set. Results from ADNI, AIBL, and OASIS-3 were therefore used to analyze the behavior and robustness of the proposed method and did not influence model selection or hyperparameter tuning.

\subsubsection{Model Design and Hyperparameter Analysis}
Table~\ref{Layer_experiments} summarizes the performance of EM\textsubscript{1} and EM\textsubscript{2} at different layers of the 3D U-Net backbone across ADNI, AIBL, and OASIS-3. Since the EM module can be integrated at different encoder depths, this analysis assessed how its placement influences generalization performance. The results demonstrate that applying the module at the second layer may provide a favorable configuration for generalization across the evaluated cohorts. Applying the module at the first layer provided moderate improvements, whereas the third layer generally degraded performance, suggesting that perturbing higher-level semantic features may disrupt discriminative information. Multi-layer perturbations showed limited benefit and, in several cases, reduced performance, particularly for simultaneous application at Layers 2 and 3.

\begin{figure}[ht]
\begin{center}

\subfloat[ADNI: $\beta_{\text{skew}}$/$\beta_{\text{kurt}}$]{%
  \includegraphics[width=0.32\textwidth]{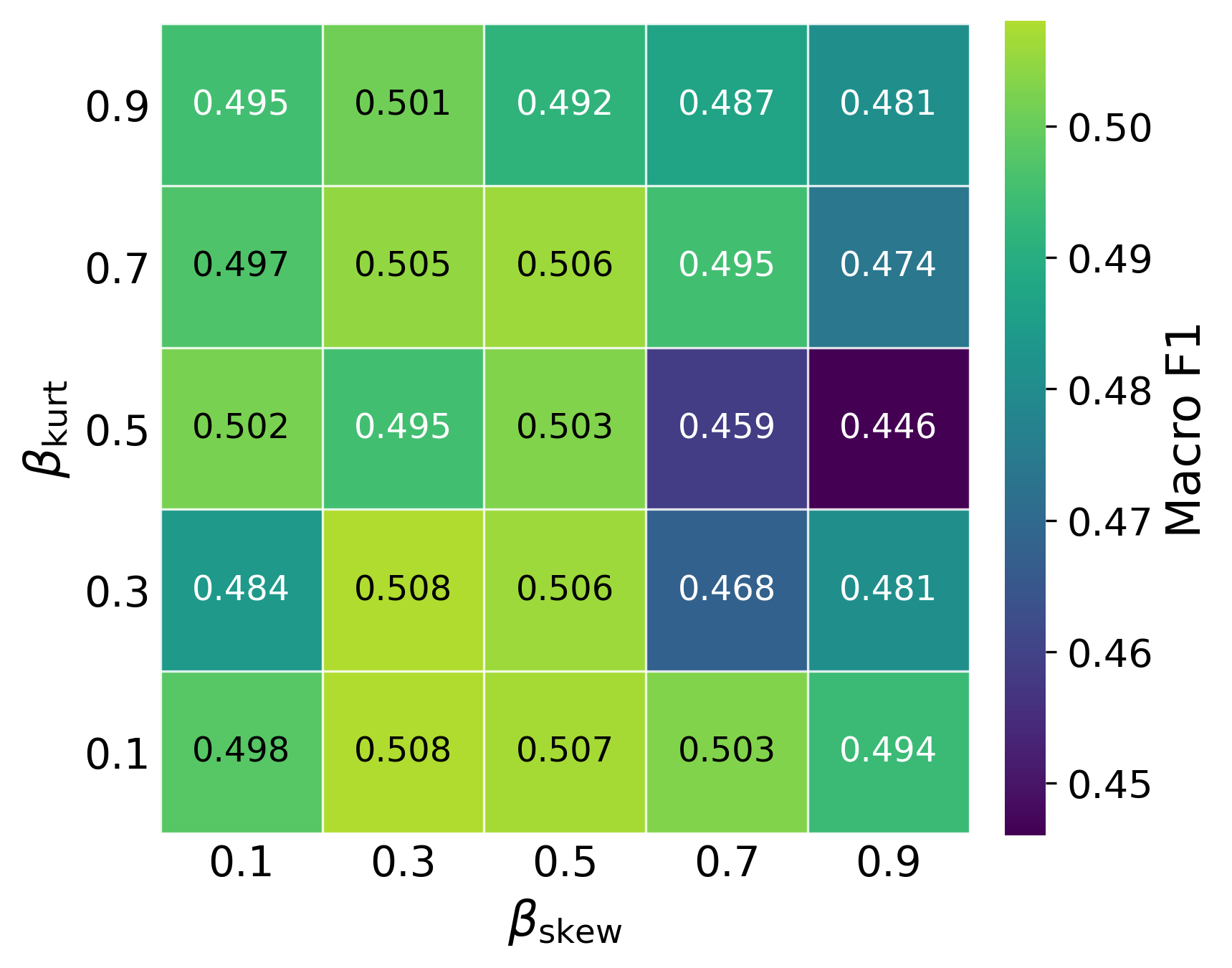}%
  \label{fig:beta_heatmap_adni}
}
\hfill
\subfloat[AIBL: $\beta_{\text{skew}}$/$\beta_{\text{kurt}}$]{%
  \includegraphics[width=0.32\textwidth]{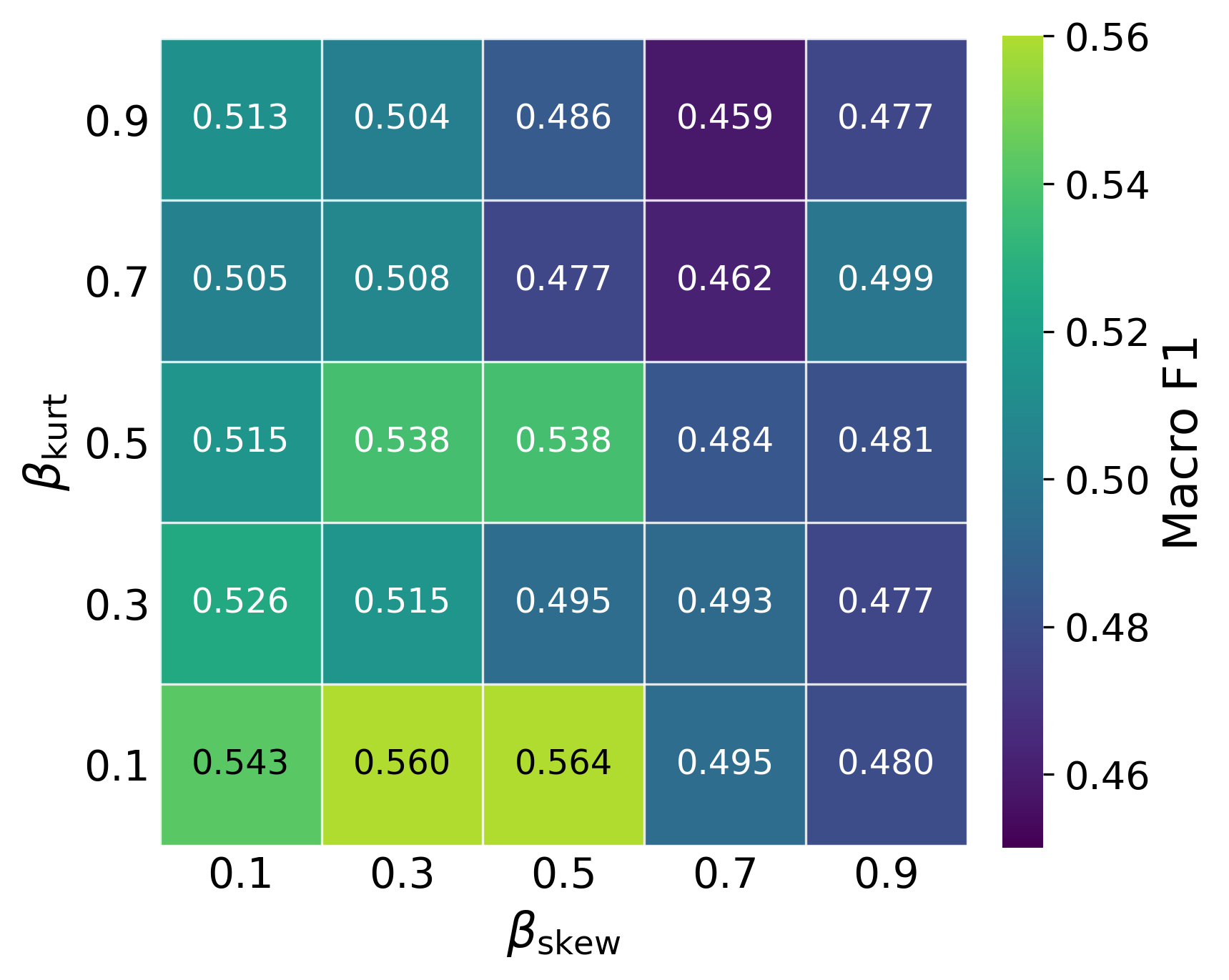}%
  \label{fig:beta_heatmap_aibl}
}
\hfill
\subfloat[OASIS-3: $\beta_{\text{skew}}$/$\beta_{\text{kurt}}$]{%
  \includegraphics[width=0.32\textwidth]{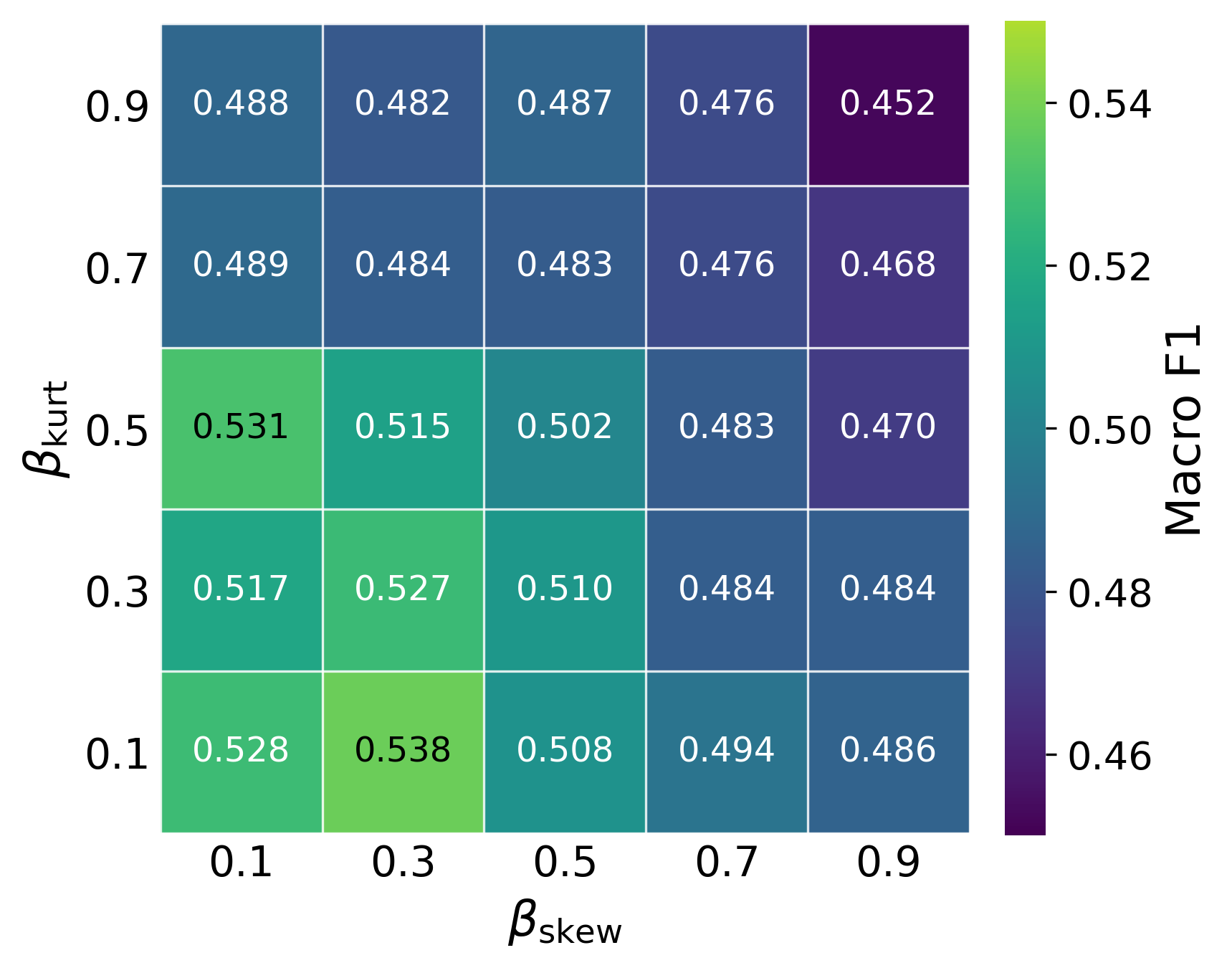}%
  \label{fig:beta_heatmap_oasis}
}

\vspace{0.3cm}

\subfloat[ADNI: $\alpha$/$p$]{%
  \includegraphics[width=0.32\textwidth]{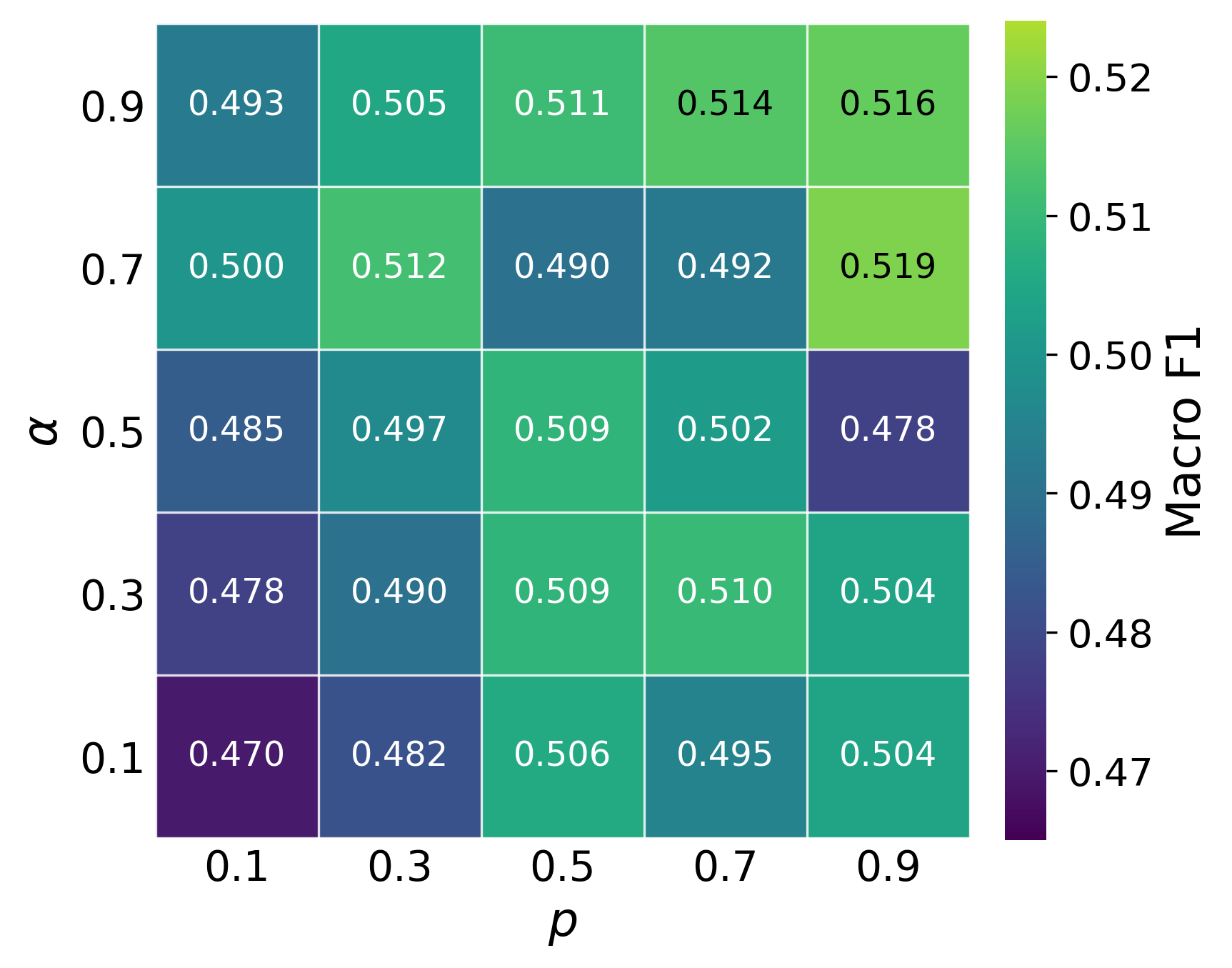}%
  \label{fig:alpha_p_heatmap_em1_adni}
}
\hfill
\subfloat[AIBL: $\alpha$/$p$]{%
  \includegraphics[width=0.32\textwidth]{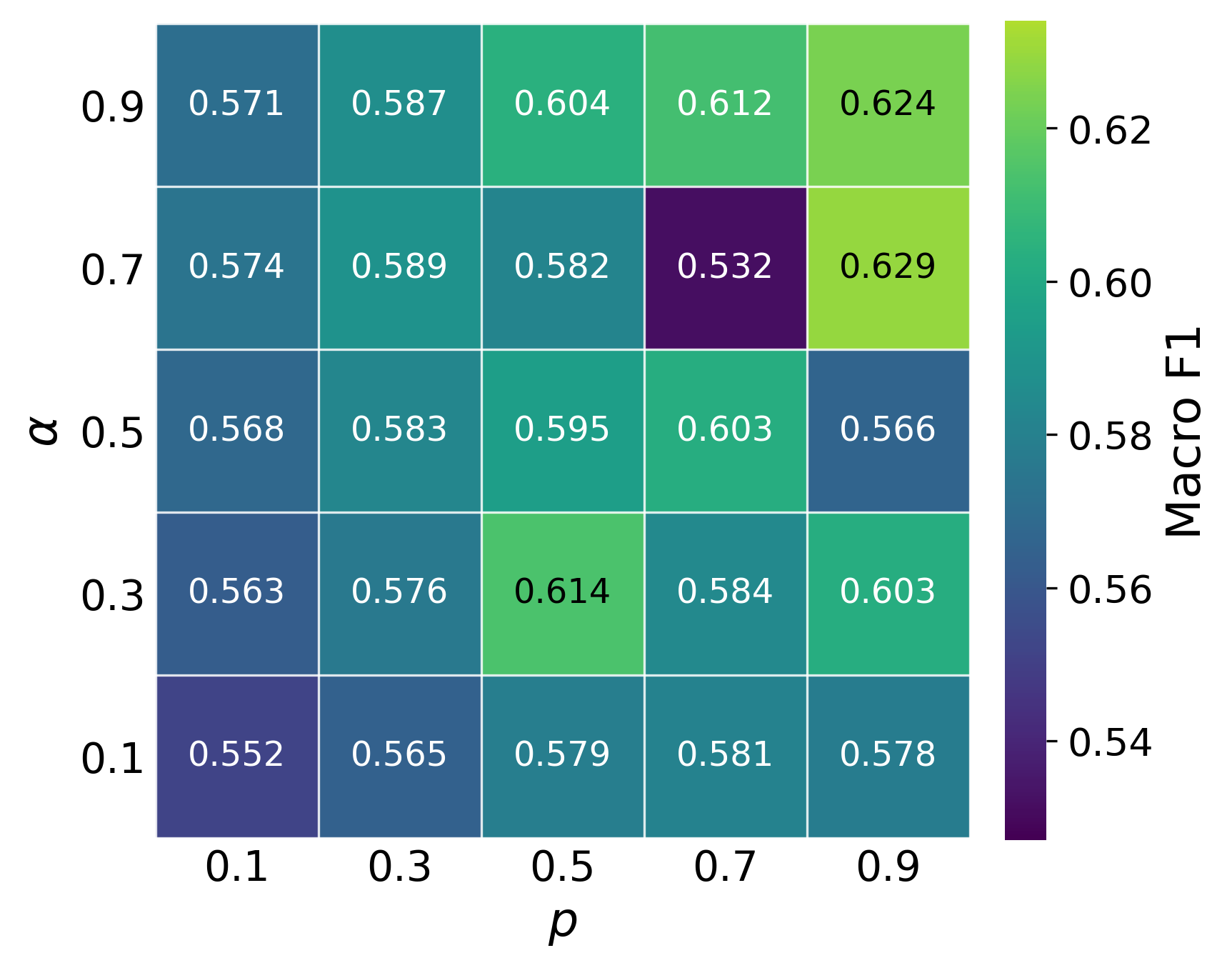}%
  \label{fig:alpha_p_heatmap_em1_aibl}
}
\hfill
\subfloat[OASIS-3: $\alpha$/$p$]{%
  \includegraphics[width=0.32\textwidth]{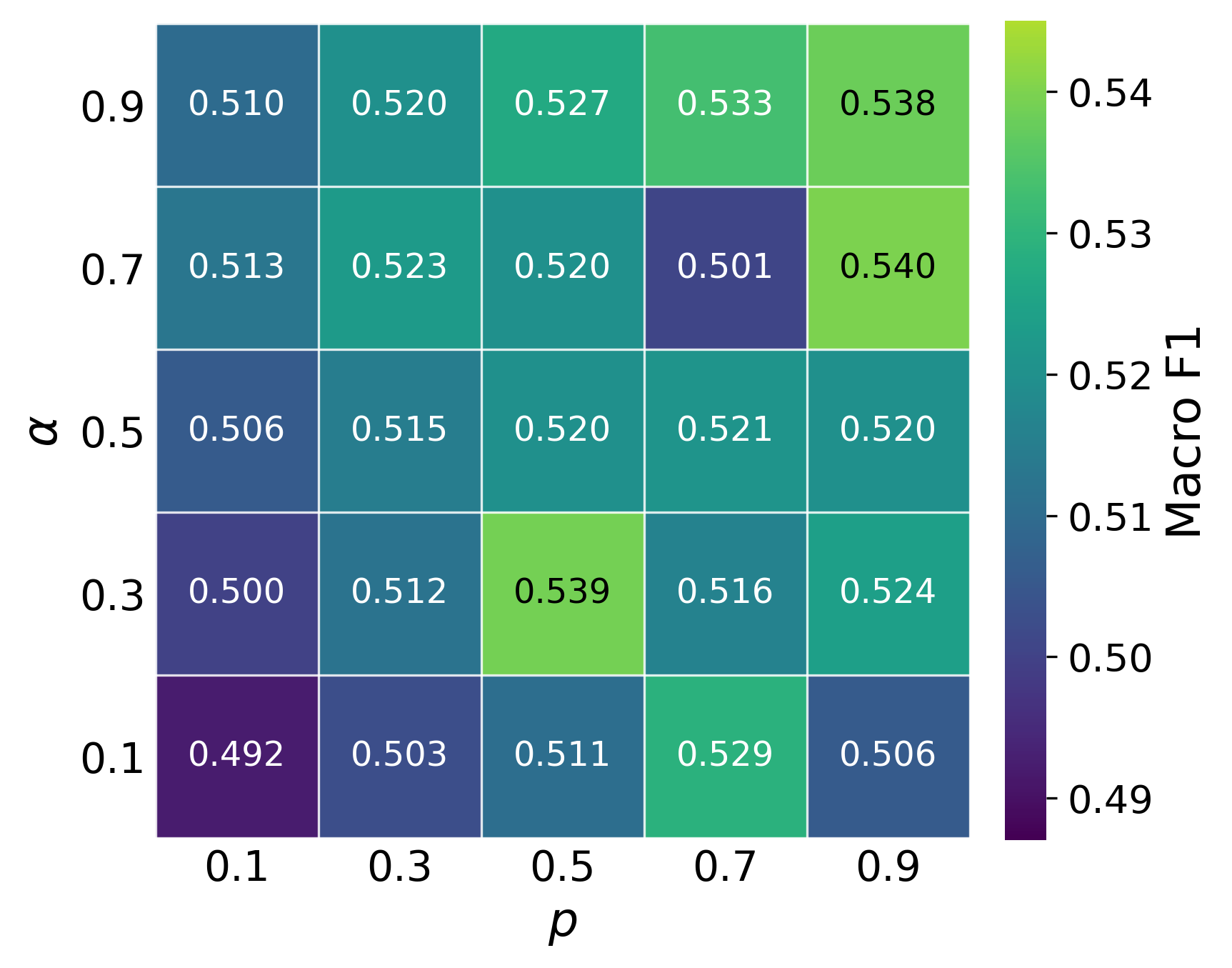}%
  \label{fig:alpha_p_heatmap_em1_oasis}
}

\caption{\textbf{Hyperparameter sensitivity analysis of Extended MixStyle.}
Top row: macro-F1 sensitivity to the higher-order weighting parameters $\beta_{\text{skew}}$ and $\beta_{\text{kurt}}$ across ADNI, AIBL, and OASIS-3. Bottom row: macro-F1 sensitivity of EM\textsubscript{1} to the MixStyle interpolation parameter $\alpha$ and mixing probability $p$. Low-to-moderate higher-order weighting generally provides more stable performance, while several neighboring $\alpha$-$p$ configurations achieve comparable results across cohorts, indicating that the selected settings are not isolated hyperparameter choices and remain robust to moderate parameter variation.}
\label{fig:hyperparameter_sensitivity}

\end{center}
\end{figure}

Fig.~\ref{fig:hyperparameter_sensitivity} summarizes the sensitivity of the proposed method to the higher-order weighting and MixStyle mixing hyperparameters. Low-to-moderate values of $\beta_{\text{skew}}$ and $\beta_{\text{kurt}}$ yield stable performance, whereas larger weights often reduce macro-F1, suggesting that excessive higher-order perturbation may distort disease-relevant features rather than simulate plausible domain shifts. The main settings, $\beta_{\text{skew}}=0.3$ and $\beta_{\text{kurt}}=0.1$, were selected using the NACC validation split and fixed for all external evaluations. For EM\textsubscript{1}, performance varies across $\alpha$ and $p$, but several neighboring configurations achieve comparable macro-F1 scores, indicating that the selected setting is not an isolated optimum. Stronger mixing generally improves ADNI and OASIS-3 performance, while AIBL shows greater fluctuation, likely due to its smaller size and class imbalance. Overall, EM\textsubscript{1} remains stable across nearby hyperparameter choices and benefits from strong feature perturbation.

\begin{table*}[ht]
\caption{\textbf{Effect of pooling strategy on cross-dataset generalization performance.} GAP, max pooling, and attention-based pooling are compared for EM\textsubscript{1} and EM\textsubscript{2}. Results show that GAP provides the most stable overall performance across external cohorts. Best results within each EM variant are in bold.}

\label{table_pooling}
\begin{center}

\setlength{\tabcolsep}{6pt}
\scriptsize
\begin{tabular}{l|cccc|cccc|cccc}

\hline

\multirow{2}{*}{\textbf{Pooling}} 

& \multicolumn{4}{c|}{\textbf{ADNI}}
& \multicolumn{4}{c|}{\textbf{AIBL}}
& \multicolumn{4}{c}{\textbf{OASIS-3}} \\

& \textbf{ACC} & \textbf{SEN} & \textbf{SPE} & \textbf{F1}
& \textbf{ACC} & \textbf{SEN} & \textbf{SPE} & \textbf{F1}
& \textbf{ACC} & \textbf{SEN} & \textbf{SPE} & \textbf{F1} \\

\hline

\multicolumn{13}{c}{\textbf{EM\textsubscript{1}}} \\
\hline

GAP
& \textbf{50.30} & 0.575 & 0.748 & \textbf{0.519}
& \textbf{76.39} & \textbf{0.614} & \textbf{0.836} & \textbf{0.629}
& \textbf{68.32} & \textbf{0.588} & \textbf{0.844} & \textbf{0.540} \\

Max
& 49.97 & \textbf{0.591} & \textbf{0.756} & 0.512
& 72.91 & 0.575 & 0.813 & 0.573
& 67.33 & 0.567 & 0.842 & 0.533 \\

Attention
& 43.05 & 0.553 & 0.731 & 0.434
& 70.95 & 0.580 & 0.814 & 0.554
& 61.58 & 0.532 & 0.838 & 0.530 \\

\hline

\multicolumn{13}{c}{\textbf{EM\textsubscript{2}}} \\
\hline

GAP
& 49.42 & \textbf{0.568} & \textbf{0.742} & \textbf{0.508}
& 66.71 & \textbf{0.613} & \textbf{0.822} & 0.560
& \textbf{64.75} & \textbf{0.601} & \textbf{0.840} & \textbf{0.538} \\

Max
& \textbf{53.26} & 0.549 & 0.738 & 0.506
& \textbf{70.49} & 0.588 & 0.818 & \textbf{0.585}
& 63.04 & 0.570 & 0.831 & 0.509 \\

Attention
& 45.52 & 0.564 & 0.739 & 0.464
& 67.47 & 0.562 & 0.811 & 0.505
& 64.47 & 0.518 & 0.834 & 0.509 \\

\hline
\end{tabular}
\end{center}
\end{table*}

To assess the effect of the pooling strategy, global average pooling (GAP), max pooling, and attention-based pooling were compared for both EM\textsubscript{1} and EM\textsubscript{2}. As shown in Table~\ref{table_pooling}, GAP provided the most stable overall performance across external cohorts. For EM\textsubscript{1}, GAP achieved the best F1 scores on ADNI, AIBL, and OASIS-3. For EM\textsubscript{2}, GAP achieved the best F1 scores on ADNI and OASIS-3, while max pooling improved the AIBL F1 score. These results suggest that GAP provides a more robust summary of high-level encoder features in the SDG setting, whereas max pooling and attention-based pooling may be more sensitive to cohort-specific activations. These post-hoc results further support the use of GAP in the source-selected final architecture.

\begin{table*}[ht]
\caption{\textbf{Fixed vs. learnable higher-order weighting parameters.} NACC-selected fixed coefficients provide more consistent overall performance, whereas learnable coefficients improve only selected cohort-specific cases. Best results within each pair are in bold.}
\label{table_learnable_params}
\begin{center}

\setlength{\tabcolsep}{5pt}
\scriptsize
\begin{tabular}{l|cccc|cccc|cccc}
\hline

\multirow{2}{*}{\textbf{Variant}} 

& \multicolumn{4}{c|}{\textbf{ADNI}}
& \multicolumn{4}{c|}{\textbf{AIBL}}
& \multicolumn{4}{c}{\textbf{OASIS-3}} \\

& \textbf{ACC} & \textbf{SEN} & \textbf{SPE} & \textbf{F1}
& \textbf{ACC} & \textbf{SEN} & \textbf{SPE} & \textbf{F1}
& \textbf{ACC} & \textbf{SEN} & \textbf{SPE} & \textbf{F1} \\

\hline

Fixed EM\textsubscript{1}

& \textbf{50.30} & 0.575 & \textbf{0.748} & \textbf{0.519}
& \textbf{76.39} & \textbf{0.614} & \textbf{0.836} & \textbf{0.629}
& 68.32 & \textbf{0.588} & \textbf{0.844} & \textbf{0.540} \\

Learnable EM\textsubscript{1}

& 49.64 & \textbf{0.578} & \textbf{0.748} & 0.510
& 76.09 & 0.571 & 0.832 & 0.585
& \textbf{71.27} & 0.581 & 0.840 & 0.533 \\

\hline

Fixed EM\textsubscript{2}

& \textbf{49.42} & \textbf{0.568} & \textbf{0.742} & \textbf{0.508}
& 66.71 & \textbf{0.613} & 0.822 & 0.560
& \textbf{64.75} & 0.601 & \textbf{0.840} & \textbf{0.538} \\

Learnable EM\textsubscript{2}

& 49.25 & 0.558 & 0.741 & 0.506
& \textbf{71.25} & 0.573 & \textbf{0.824} & \textbf{0.581}
& 61.49 & \textbf{0.625} & 0.836 & 0.519 \\

\hline
\end{tabular}

\end{center}
\end{table*}

Moreover, to examine whether higher-order statistics should be fixed or adaptively weighted, we evaluated learnable skewness and kurtosis weights. Table~\ref{table_learnable_params} shows cohort-specific gains, especially for EM\textsubscript{2} on AIBL, where F1 increased from 0.560 to 0.581. However, the improvement was inconsistent: fixed weighting performed better overall for EM\textsubscript{1}, while for EM\textsubscript{2}, learnable weighting improved AIBL but reduced OASIS-3 and slightly reduced ADNI performance. Thus, adaptive weighting is promising but may require additional regularization, so fixed weighting was retained for stability and reproducibility.

\subsubsection{Representation and Diagnostic Analysis}

\begin{figure}[ht]
  \begin{center}
  \textbf{ADNI} \\[-0.5ex]
  \subfloat[Original sMRI]{%
    \includegraphics[height=3cm]{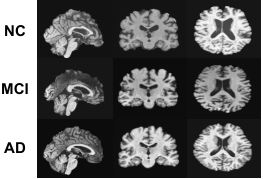}
    \label{fig:gradviz-adni-input}
  }%
  \subfloat[MixStyle]{%
    \includegraphics[height=3cm]{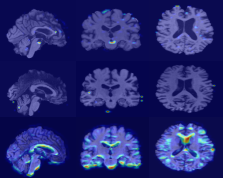}
    \label{fig:gradviz-adni-nomixstyle}
  }%
  \subfloat[EM\textsubscript{1}]{%
    \includegraphics[height=3cm]{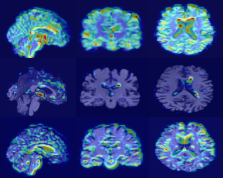}
    \label{fig:gradviz-adni-meanvar}
  }%
  \subfloat[EM\textsubscript{2}]{%
    \includegraphics[height=3cm]{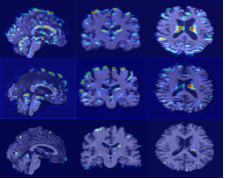}
    \label{fig:gradviz-adni-extended}
  }\\[2ex]

  \textbf{AIBL} \\[-0.5ex]
  \subfloat[Original sMRI]{%
    \includegraphics[height=3cm]{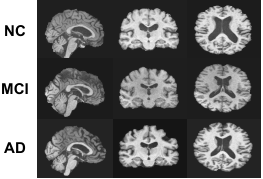}
    \label{fig:gradviz-aibl-input}
  }%
  \subfloat[MixStyle]{%
    \includegraphics[height=3cm]{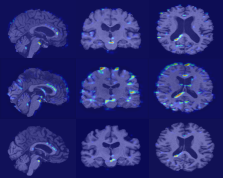}
    \label{fig:gradviz-aibl-nomixstyle}
  }%
  \subfloat[EM\textsubscript{1}]{%
    \includegraphics[height=3cm]{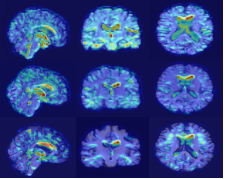}
    \label{fig:gradviz-aibl-meanvar}
  }%
  \subfloat[EM\textsubscript{2}]{%
    \includegraphics[height=3cm]{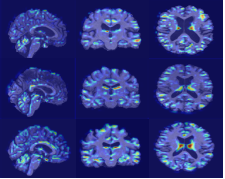}
    \label{fig:gradviz-aibl-extended}
  }\\[2ex]

  \textbf{OASIS-3} \\[-0.5ex]
  \subfloat[Original sMRI]{%
    \includegraphics[height=3cm]{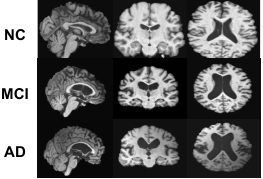}
    \label{fig:gradviz-oasis-input}
  }%
  \subfloat[MixStyle]{%
    \includegraphics[height=3cm]{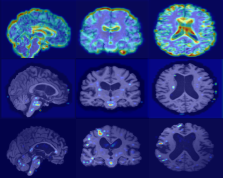}
    \label{fig:gradviz-oasis-nomixstyle}
  }%
  \subfloat[EM\textsubscript{1}]{%
    \includegraphics[height=3cm]{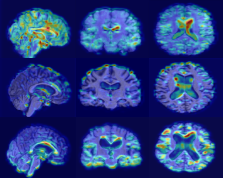}
    \label{fig:gradviz-oasis-meanvar}
  }%
  \subfloat[EM\textsubscript{2}]{%
    \includegraphics[height=3cm]{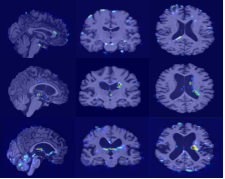}
    \label{fig:gradviz-oasis-extended}
  }

  \caption{\textbf{Grad-CAM visualizations on 3D sMRI samples across cohorts.} The figure presents NC, MCI and AD subjects from ADNI (top row), AIBL (middle row), and OASIS-3 (bottom row). For each cohort, columns show: original sMRI scans, MixStyle baseline, EM\textsubscript{1} based on mean, standard deviation, and skewness, and EM\textsubscript{2} extending EM\textsubscript{1} with kurtosis.
}
  \label{fig:gradcam_combined}
  \end{center}
\end{figure}

Grad-CAM visualizations in Fig.~\ref{fig:gradcam_combined} show that, compared to MixStyle, the proposed EM variants produce more stable and focused activations within cortical and subcortical regions commonly affected by AD. EM\textsubscript{1} provides the clearest localization, reducing noisy responses outside brain tissue and highlighting disease-relevant areas more consistently. EM\textsubscript{2} shows a similar trend, though its attention maps are slightly more diffuse.

\begin{figure}[ht]
  \begin{center}
  \subfloat[3D U-Net]{%
    \includegraphics[width=0.47\textwidth]{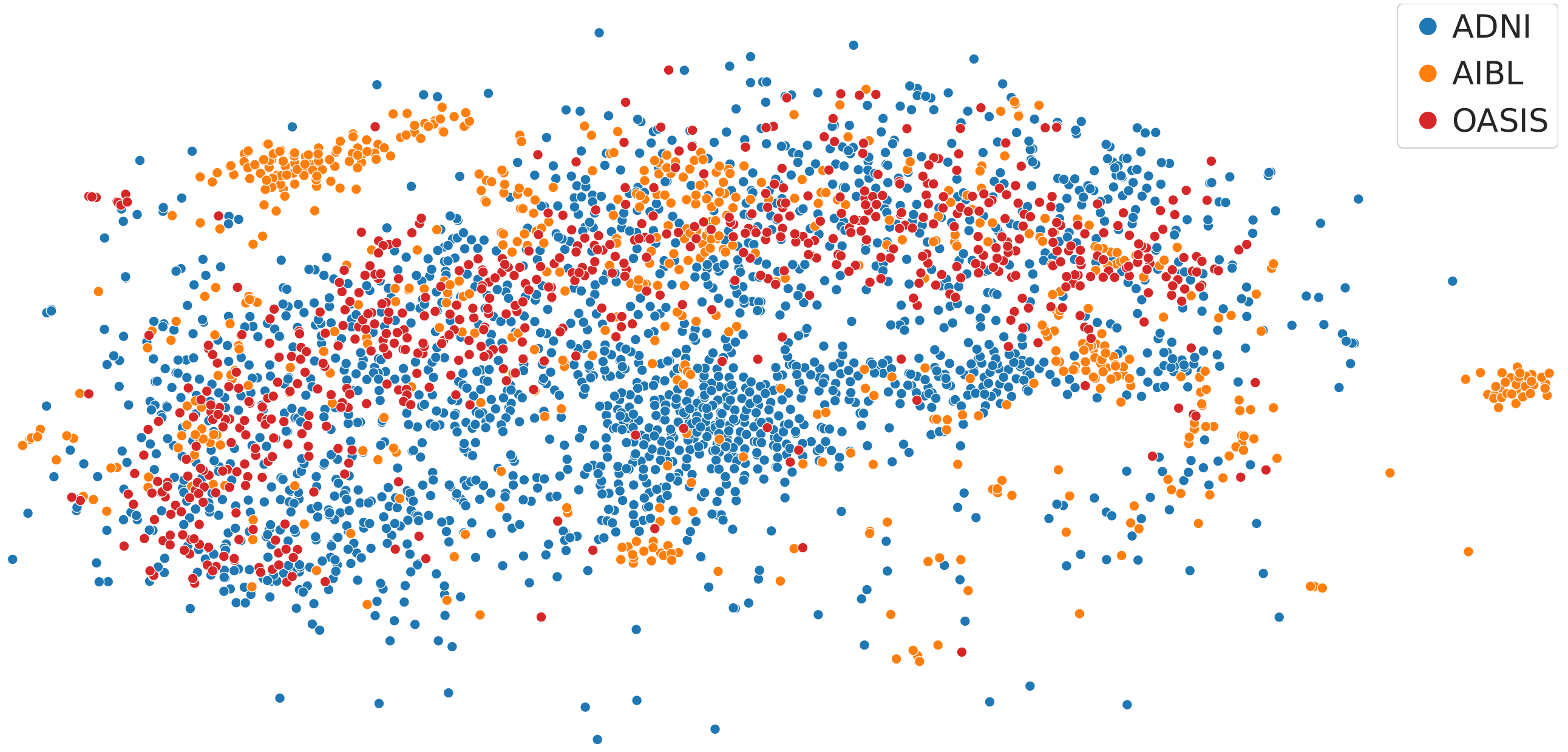}%
    \label{fig:tsne-1}
  }\hfill
  \subfloat[EFDM]{%
    \includegraphics[width=0.47\textwidth]{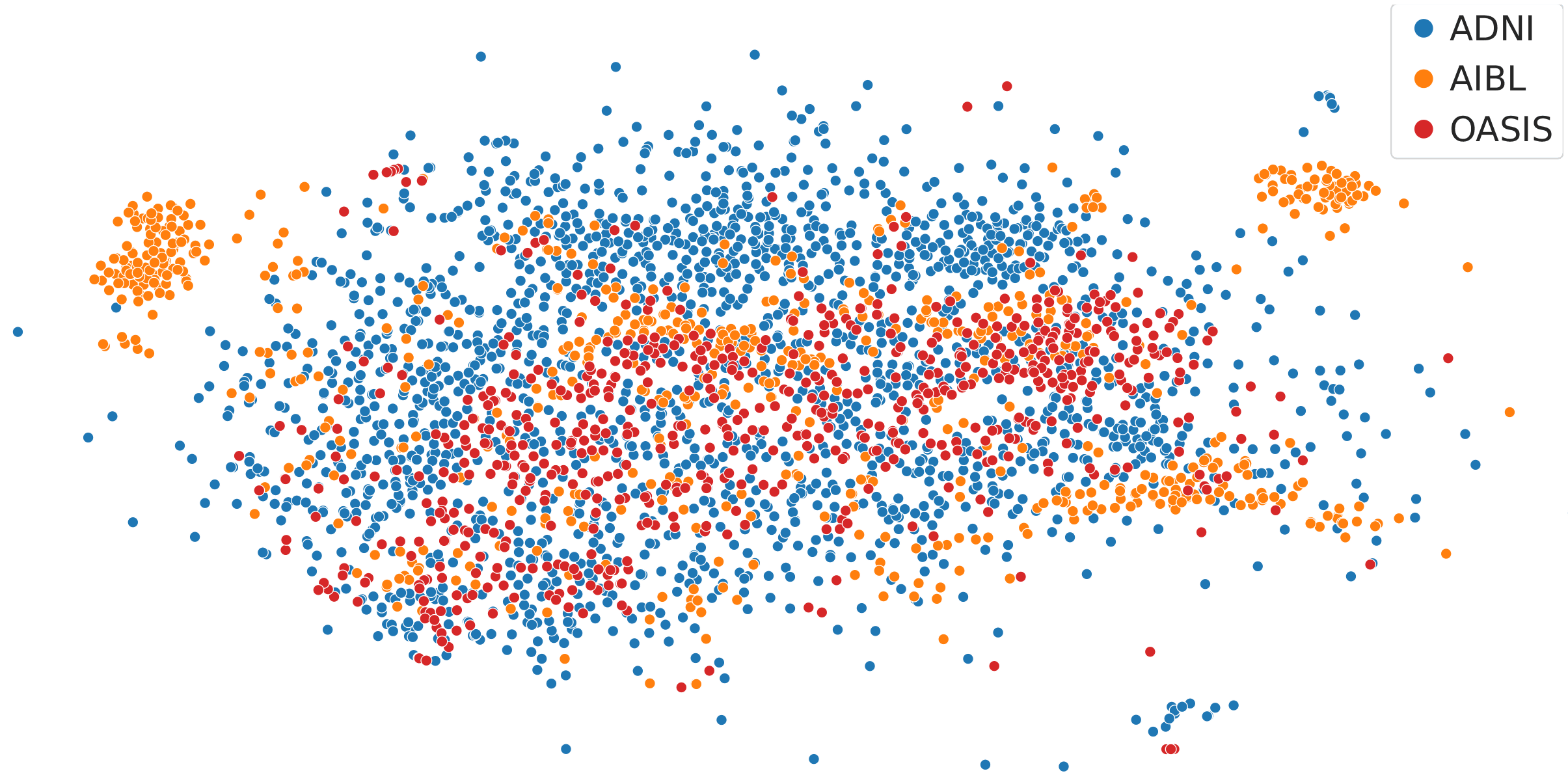}%
    \label{fig:tsne-2}
  }

  \par\bigskip 

  \subfloat[EM\textsubscript{1}]{%
    \includegraphics[width=0.47\textwidth]{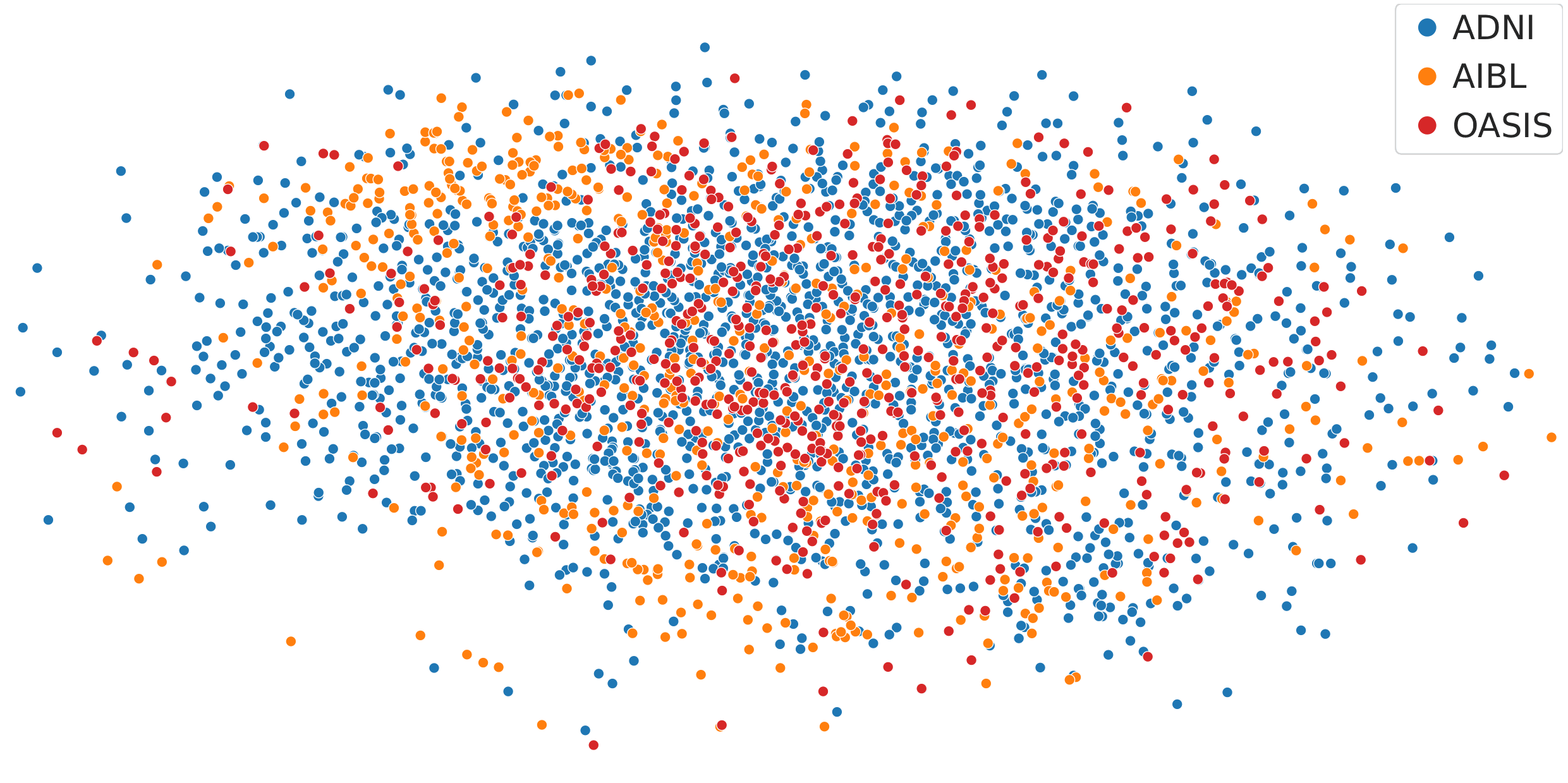}%
    \label{fig:tsne-3}
  }\hfill
  \subfloat[EM\textsubscript{2}]{%
    \includegraphics[width=0.47\textwidth]{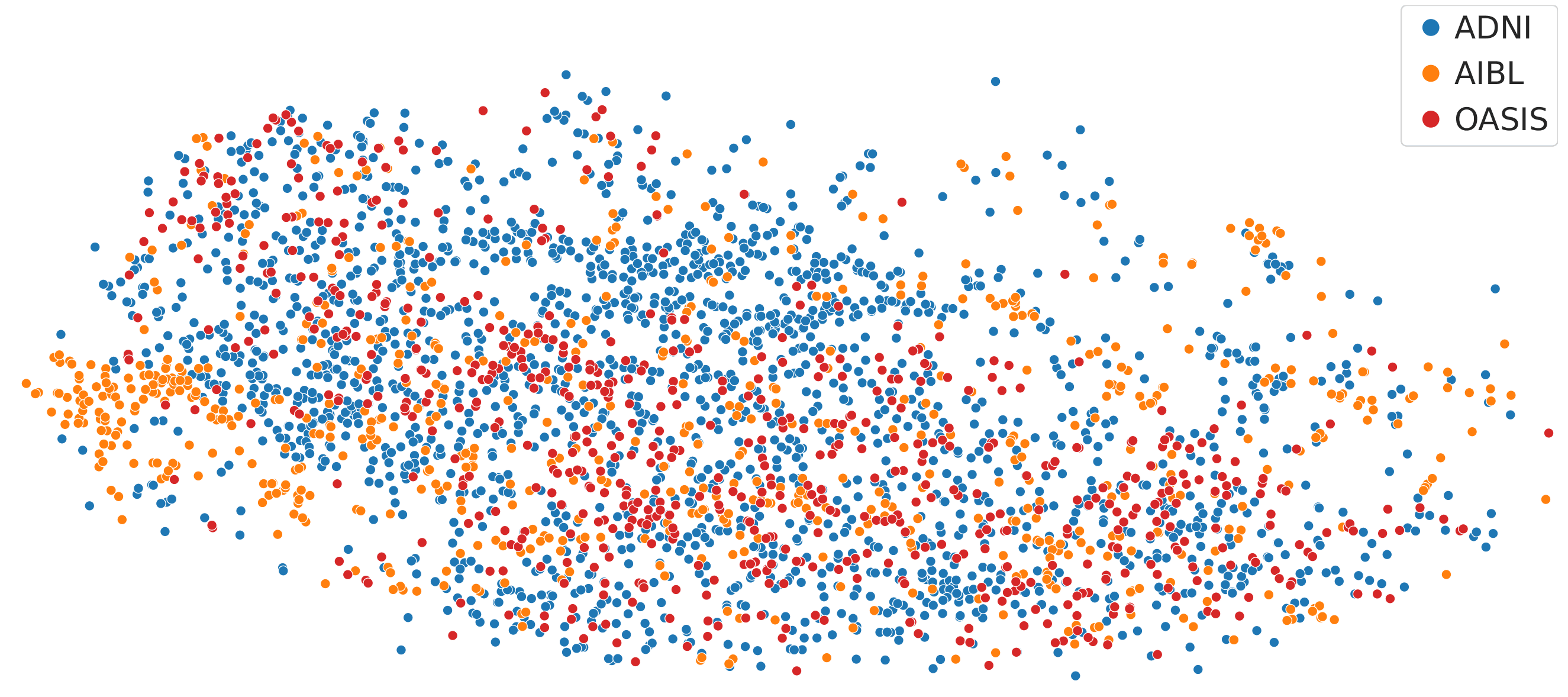}%
    \label{fig:tsne-4}
  }

\caption{\textbf{t-SNE visualizations of sMRI embeddings under different training settings.} Data were drawn from three cohorts: ADNI, AIBL and OASIS-3. The vanilla 3D U-Net (a) shows more cohort-specific separation, with AIBL forming relatively compact clusters. EFDM (b) increases inter-cohort overlap. EM\textsubscript{1} (c) shows the strongest intermixing among the external cohorts, while EM\textsubscript{2} (d) also improves overlap relative to the baseline, though with somewhat less uniform mixing.}

  \label{fig:tsne_combined}
  \end{center}
\end{figure}

Across the t-SNE embeddings in Fig.~\ref{fig:tsne_combined}, since NACC was used as the source training cohort, only the external target cohorts were visualized to assess alignment among unseen domains. The 3D U-Net in Fig.~\ref{fig:tsne-1} showed the clearest cohort separation, with AIBL forming several islands and OASIS-3 concentrated in the upper region while EFDM in Fig.~\ref{fig:tsne-2} improved inter-cohort mixing. EM\textsubscript{1} in Fig.~\ref{fig:tsne-3} further dispersed cohort-specific islands, distributing AIBL and OASIS-3 more uniformly and increasing overlap throughout the embedding without obvious isolated clusters. EM\textsubscript{2} in Fig.~\ref{fig:tsne-4} shows a similar degree of mixing to EM\textsubscript{1}, with a slightly tighter interleaved core and only a few outer regions dominated by individual target cohorts. Overall, the progression from baseline to EFDM and then to EM\textsubscript{1}/EM\textsubscript{2} illustrates a shift from dataset-driven clustering toward reduced cohort bias.

\begin{table*}[!ht]
\caption{\textbf{One-to-all (AD vs. all) analysis to evaluate AD classification performance across cohorts.} Performance is compared among the 3D U-Net baseline, EFDM (best performing competing method), and the proposed EM$_1$ and EM$_2$ variants. Best results are in bold.}

\label{table_one_to_all}
\begin{center}
\setlength{\tabcolsep}{6pt} 
\scriptsize
\begin{tabular}{l|cccc|cccc|cccc}
\hline
\multirow{2}{*}{\textbf{Methods}} & \multicolumn{4}{c|}{\textbf{ADNI}} & \multicolumn{4}{c|}{\textbf{AIBL}} & \multicolumn{4}{c}{\textbf{OASIS-3}} \\
 & \textbf{ACC} & \textbf{SEN} & \textbf{SPE} & \textbf{F1} 
 & \textbf{ACC} & \textbf{SEN} & \textbf{SPE} & \textbf{F1} 
 & \textbf{ACC} & \textbf{SEN} & \textbf{SPE} & \textbf{F1} \\
\hline
Baseline \cite{Unet3D}  & 81.16 & 0.626 & 0.859 & 0.572  
           & 91.53 & \textbf{0.620} & 0.955 & 0.636  
           & \textbf{84.16} & \textbf{0.570} & 0.958 & \textbf{0.683} \\
EFDM \cite{zhang2022exact}      & 78.29 & 0.558 & 0.840 & 0.510  
           & 90.92 & 0.468 & 0.969 & 0.552  
           & 82.29 & 0.518 & 0.953 & 0.637 \\
EM\textsubscript{1} & \textbf{83.56} & \textbf{0.645} & \textbf{0.884} & \textbf{0.614}  
           & 91.51 & 0.481 & \textbf{0.974} & 0.575 
           & 81.37 & 0.451 & \textbf{0.969} & 0.592 \\
EM\textsubscript{2} & 81.43 & 0.615 & 0.865 & 0.573  
           & \textbf{92.32} & 0.608 & 0.966 & \textbf{0.654}  
           & 84.01 & 0.554 & 0.962 & 0.675 \\
\hline
\end{tabular}
\end{center}
\end{table*}

Table~\ref{table_one_to_all} presents the one-to-all AD detection results. Although the main task is multiclass classification, this evaluation examines the ability of the learned representations to distinguish AD from the remaining diagnostic groups. EM\textsubscript{1} achieved the highest F1 score on ADNI, improving over EFDM by 10.4 percentage points, while EM\textsubscript{2} achieved the highest F1 score on AIBL, improving over EFDM by 10.2 percentage points. On OASIS-3, the baseline achieved the highest F1 score, with EM\textsubscript{2} comparable and 3.8 percentage points above EFDM. Overall, the benefits of higher-order perturbations vary across cohorts, and neither variant consistently improves performance across datasets.

\begin{table*}[ht]
\caption{\textbf{Binary classification performance for clinically relevant cross-dataset diagnostic tasks.} Classification results are reported for MCI vs. NC and MCI vs. AD on the external ADNI, AIBL, and OASIS-3 cohorts. The baseline 3D U-Net is compared with EM\textsubscript{1} and EM\textsubscript{2}. Best results within each binary task and cohort are in bold.}
\label{table_binary_tasks}
\begin{center}

\setlength{\tabcolsep}{4.4pt}
\scriptsize
\begin{tabular}{ll|cccc|cccc|cccc}
\hline

\multirow{2}{*}{\textbf{Task}} &
\multirow{2}{*}{\textbf{Method}}

& \multicolumn{4}{c|}{\textbf{ADNI}}
& \multicolumn{4}{c|}{\textbf{AIBL}}
& \multicolumn{4}{c}{\textbf{OASIS-3}} \\

&
& \textbf{ACC} & \textbf{SEN} & \textbf{SPE} & \textbf{F1}
& \textbf{ACC} & \textbf{SEN} & \textbf{SPE} & \textbf{F1}
& \textbf{ACC} & \textbf{SEN} & \textbf{SPE} & \textbf{F1} \\

\hline

\multirow{3}{*}{MCI vs NC}

& Baseline
& 47.90 & 0.374 & \textbf{0.691} & 0.490
& \textbf{71.30} & 0.481 & \textbf{0.762} & 0.369
& 63.09 & 0.629 & 0.631 & 0.170 \\

& EM\textsubscript{1} 
& \textbf{56.81} & \textbf{0.565} & 0.573 & \textbf{0.636}
& 66.76 & \textbf{0.647} & 0.672 & \textbf{0.406}
& 61.41 & 0.666 & 0.611 & 0.172 \\

& EM\textsubscript{2} 
& 56.40 & 0.518 & 0.656 & 0.613
& 69.24 & 0.529 & 0.727 & 0.376
& \textbf{67.18} & \textbf{0.740} & \textbf{0.667} & \textbf{0.212} \\

\hline

\multirow{3}{*}{MCI vs AD}

& Baseline
& 67.38 & 0.739 & 0.649 & 0.556
& 70.65 & 0.556 & \textbf{0.823} & 0.624
& 63.68 & 0.626 & 0.714 & 0.752 \\

& EM\textsubscript{1} 
& \textbf{68.95} & \textbf{0.796} & 0.649 & \textbf{0.586}
& \textbf{70.71} & \textbf{0.683} & 0.725 & \textbf{0.671}
& 65.84 & \textbf{0.647} & 0.740 & \textbf{0.769} \\

& EM\textsubscript{2} 
& 68.58 & 0.772 & \textbf{0.653} & 0.575
& 69.52 & 0.556 & 0.803 & 0.615
& \textbf{65.90} & 0.642 & \textbf{0.777} & 0.767 \\

\hline
\end{tabular}

\end{center}
\end{table*}

\begin{table*}[ht]
\caption{\textbf{Cross-source domain generalization experiments.} Models are trained on one source cohort and evaluated on the remaining target cohorts. For each source setting, the baseline 3D U-Net is compared with EM\textsubscript{1} and EM\textsubscript{2}.  Best results are in bold.}
\label{table_cross_source_v3}
\begin{center}

\setlength{\tabcolsep}{2pt}
\scriptsize
\begin{tabular}{l|cccc|cccc|cccc|cccc}
\hline

\textbf{Method} &

\multicolumn{4}{c|}{\textbf{ADNI}} &
\multicolumn{4}{c|}{\textbf{AIBL}} &
\multicolumn{4}{c|}{\textbf{OASIS-3}} &
\multicolumn{4}{c}{\textbf{NACC}} \\

& \textbf{ACC} & \textbf{SEN} & \textbf{SPE} & \textbf{F1}
& \textbf{ACC} & \textbf{SEN} & \textbf{SPE} & \textbf{F1}
& \textbf{ACC} & \textbf{SEN} & \textbf{SPE} & \textbf{F1}
& \textbf{ACC} & \textbf{SEN} & \textbf{SPE} & \textbf{F1} \\

\hline

\multicolumn{17}{c}{\textbf{Source Dataset: ADNI}} \\
\hline

Baseline 
& - & - & - & -
& 48.26 & 0.497 & 0.764 & 0.453
& 49.22 & 0.494 & 0.782 & 0.425
& 54.36 & 0.538 & 0.774 & 0.535 \\

EM\textsubscript{1}
& - & - & - & -
& \textbf{58.69} & 0.529 & \textbf{0.792} & 0.509
& \textbf{55.43} & \textbf{0.561} & \textbf{0.814} & \textbf{0.476}
& \textbf{56.69} & \textbf{0.547} & \textbf{0.782} & \textbf{0.548} \\

EM\textsubscript{2}
& - & - & - & -
& 57.33 & \textbf{0.562} & 0.788 & \textbf{0.534}
& 50.77 & 0.526 & 0.793 & 0.442
& 55.70 & 0.545 & 0.778 & \textbf{0.548} \\

\hline

\multicolumn{17}{c}{\textbf{Source Dataset: AIBL}} \\
\hline

Baseline
& 31.74 & \textbf{0.424} & 0.693 & 0.274
& - & - & - & -
& 68.78 & 0.445 & 0.758 & 0.448
& 58.82 & 0.431 & 0.720 & 0.402 \\

EM\textsubscript{1}
& \textbf{33.71} & 0.422 & \textbf{0.695} & \textbf{0.301}
& - & - & - & -
& 69.87 & \textbf{0.456} & 0.765 & \textbf{0.455}
& \textbf{60.82} & 0.454 & \textbf{0.734} & \textbf{0.427} \\

EM\textsubscript{2}
& 32.07 & 0.401 & 0.688 & 0.281
& - & - & - & -
& \textbf{73.44} & 0.445 & \textbf{0.768} & 0.445
& 58.06 & \textbf{0.460} & 0.733 & 0.411 \\

\hline

\multicolumn{17}{c}{\textbf{Source Dataset: OASIS-3}} \\
\hline

Baseline
& 34.10 & 0.466 & 0.705 & 0.308
& 72.01 & 0.501 & 0.786 & 0.443
& - & - & - & -
& 59.66 & 0.462 & 0.736 & 0.410 \\

EM\textsubscript{1} 
& \textbf{36.51} & \textbf{0.507} & \textbf{0.718} & \textbf{0.331}
& \textbf{74.73} & \textbf{0.534} & \textbf{0.794} & \textbf{0.453}
& - & - & - & -
& \textbf{61.89} & \textbf{0.517} & \textbf{0.768} & \textbf{0.440} \\

EM\textsubscript{2} 
& 35.85 & 0.497 & 0.714 & 0.327
& 73.52 & 0.527 & 0.764 & \textbf{0.453}
& - & - & - & -
& 61.76 & 0.502 & 0.758 & 0.435 \\

\hline
\end{tabular}

\end{center}
\end{table*}

To further assess clinically relevant distinctions, binary cross-dataset evaluations were conducted for MCI vs. NC and MCI vs. AD. As shown in Table~\ref{table_binary_tasks}, EM\textsubscript{1} achieved the highest F1 score across all external cohorts for both tasks, except for MCI vs. NC on OASIS-3 (where EM\textsubscript{2} was highest), while EM\textsubscript{2} provided the highest OASIS-3 sensitivity/specificity for MCI vs. NC and the highest ADNI/OASIS-3 specificity for MCI vs. AD. The OASIS-3 findings should be interpreted in the context of its limited MCI representation (27 participants), which may contribute to greater variability in the corresponding binary performance estimates. These results
suggest potential benefits for binary classification, although
the observed numerical differences vary across cohorts and
should be interpreted cautiously.

\subsubsection{Generalization and Computational Analysis}

In an additional cross-source analysis, to evaluate cross-dataset generalization beyond the NACC-source setting, additional experiments were conducted using ADNI, AIBL, and OASIS-3 as source cohorts. For each source cohort, the baseline 3D U-Net, EM\textsubscript{1}, and EM\textsubscript{2} were trained on the source dataset and evaluated on the remaining external cohorts, including NACC. As shown in Table~\ref{table_cross_source_v3}, EM\textsubscript{1} achieved numerically higher performance than the baseline in most source-target settings, suggesting that the proposed higher-order perturbation strategy is not restricted to NACC as the source cohort.

\begin{table}[ht]
\caption{\textbf{Computational efficiency comparison of the baseline and augmentation-based methods.} Training time is reported as mean $\pm$ standard deviation over three measured epochs.}

\begin{center}

\setlength{\tabcolsep}{6pt}
\scriptsize
\begin{tabular}{l|c c c}
\hline
\textbf{Method} & \textbf{Training Time} & \textbf{Overhead} & \textbf{Peak GPU} \\
                & \textbf{/Epoch (s)}    & \textbf{(\%)}    & \textbf{Memory (GB)} \\
\hline
Baseline \cite{Unet3D}            & $659.62 \pm 2.82$  & 0.00  & 8.69 \\
MixUp \cite{zhang2018mixup}       & $759.62 \pm 2.82$  & 15.16 & 8.76 \\
MixStyle \cite{mixstyle}          & $805.19 \pm 14.42$ & 22.07 & 8.88 \\
EM\textsubscript{1}               & $856.96 \pm 4.30$  & 29.92 & 9.10 \\
EM\textsubscript{2}               & $875.73 \pm 8.40$  & 32.76 & 9.58 \\
\hline
\end{tabular}
\end{center}
\label{tab:computational_efficiency}
\end{table}

Lastly, Table~\ref{tab:computational_efficiency} compares training-time efficiency across methods on the NACC dataset. EM\textsubscript{1} and EM\textsubscript{2} introduce moderate overheads in per-epoch training time (29.92\% and 32.76\%, respectively) and peak GPU memory relative to the baseline 3D U-Net. Since the EM modules are applied only during training, these additional costs do not affect the inference-time architecture or parameter count.

\section{Conclusion}
In this work, we extended the MixStyle framework by incorporating higher-order statistics into feature normalization to improve domain generalization in classifying cognitive decline phenotypes from 3D sMRI. Evaluations on ADNI, AIBL, and OASIS-3 showed that the skewness-only variant achieved the most consistent overall performance among the evaluated methods, with statistically significant improvements over 3D U-Net on ADNI, despite class imbalance and protocol variability. This variant required moderately more training time but added no inference-time parameters or model complexity. Future work will focus on computational efficiency and validation in larger, diverse prospective cohorts to assess clinical applicability.

\bibliographystyle{IEEEtran}
\bibliography{references}
\end{document}